%% file: main-cam.tex
\documentclass[sigconf]{acmart}

\usepackage{amsfonts}
\usepackage{multirow}
\usepackage{amsmath}
\usepackage{xcolor}
\usepackage{xspace}
\usepackage{algorithm} 
\usepackage{algpseudocode} 

\usepackage{subcaption}

\newcommand{\ie}{\emph{i.e.},\xspace}
\newcommand{\eg}{\emph{e.g.},\xspace}

\newcommand{\etc}{\emph{etc}\xspace}

\def\B#1{\mathbf #1}
\def\C#1{\mathcal #1}

\AtBeginDocument{%
	}

\setcopyright{acmcopyright}
\copyrightyear{2023}
\acmYear{2023}
\acmDOI{XXXXXXX.XXXXXXX}

\acmConference[WSDM '23]{WSDM '23: The 16th ACM International Conference on Web Search and Data Mining}{February 27--March 3, 2023}{Singapore}
\acmBooktitle{WSDM '23: The 16th ACM International Conference on Web Search and Data Mining, February 27--March 3, 2023, Singapore}
\acmPrice{15.00}
\acmISBN{978-1-4503-XXXX-X/18/06}

\begin{document}
\title{Inductive Graph Transformer for Delivery Time Estimation}

\author{
	Xin Zhou$^1$, Jinglong Wang$^2$, 
	Yong Liu$^1$, 
	Xingyu Wu$^2$, 
	Zhiqi Shen$^1$,
	Cyril Leung$^1$
}

\affiliation{\vspace{0.1cm}
	$^1$ Nanyang Technological University \country{Singapore}  \qquad\qquad\qquad\qquad   $^2$  Alibaba Group \country{China}\qquad\qquad\qquad\quad \\
	\{xin.zhou, stephenliu, zqshen, cleung\}@ntu.edu.sg \quad \quad \{jinglong.wjl, zhuyang.wxy\}@alibaba-inc.com 
}

\renewcommand{\shortauthors}{Xin Zhou, Jinglong Wang, Yong Liu, Xingyu Wu, Zhiqi Shen, and Cyril Leung}

\begin{abstract}
	Providing accurate estimated time of package delivery on users' purchasing pages for e-commerce platforms is of great importance to their purchasing decisions and post-purchase experiences. 
	Although this problem shares some common issues with the conventional estimated time of arrival (ETA), it is more challenging with the following aspects:
	1) \textit{Inductive inference.} 
	Models are required to predict ETA for orders with unseen retailers and addresses;
	2) \textit{High-order interaction of order semantic information.} 
	Apart from the spatio-temporal features, the estimated time also varies greatly with other factors, such as the packaging efficiency of retailers, as well as the high-order interaction of these factors.
	In this paper, we propose an inductive graph transformer (IGT) that leverages raw feature information and structural graph data to estimate package delivery time.
	Different from previous graph transformer architectures, IGT adopts a decoupled pipeline and trains transformer as a regression function that can capture the multiplex information from both raw feature and dense embeddings encoded by a graph neural network (GNN).
	In addition, we further simplify the GNN structure by removing its non-linear activation and the learnable linear transformation matrix. The reduced parameter search space and linear information propagation in the simplified GNN enable the IGT to be applied in large-scale industrial scenarios.
	Experiments on real-world logistics datasets show that our proposed model can significantly outperform the state-of-the-art methods on estimation of delivery time. The source code is available at: https://github.com/enoche/IGT-WSDM23.
\end{abstract}



\maketitle

\input{content-cam.tex}

\begin{acks}
	This research is supported, in part, by Alibaba Group through Alibaba Innovative Research (AIR) Program and Alibaba-NTU Singapore Joint Research Institute (JRI)(No. AN-GC-2020-019).
\end{acks}

\bibliographystyle{ACM-Reference-Format}
\bibliography{main-cam}

	
\end{document}

%% file: content-cam.tex
\section{Introduction}

The invention of the World Wide Web enables users to share information and products on the Web, which eventually drives the development of e-commerce platforms~\cite{vladimir1996electronic}.
The contemporary prevalence of e-commerce leads to millions of packages being delivered worldwide every day~\cite{doole2020estimation, su2020characterizing}.
Retailers on e-commerce platforms show consumers various information about their products and services, such as prices, descriptions, and estimated delivery time. 
The delivery time is valued more important than price when consumers have a deadline to receive their packages. It has been shown to affect the decisions of 87\% online consumers~\cite{cui2020sooner}.
Hence, the accurate estimation of package arrival time (ETA) not only impacts the e-commerce revenue but also shapes the consumers' expectations. 
Either over-promising or under-promising ETA may result in detriment of online retailers and consumers.

In practice, the physical delivery time is always subject to uncertainty due to the packaging efficiency of retailers, inventory, scheduling, or transportation factors. 
In our scenario, we study the origin-destination (OD) delivery time estimation, which aims to predict the package delivery time given the information of retailer, OD addresses, and payment time.
Although the OD travel time estimation is widely studied in the transportation domain~\cite{li2018multi, hu2020stochastic}, its challenges are usually different from those of OD ETA in the delivery scenario. 
For example, both problems face the challenges of the absence of route information and efficiently utilizing the spatio-temporal features. 
However, the OD ETA in the delivery scenario inevitably involves predicting orders with unseen retailers or addresses, \etc. 
Furthermore, its prediction varies greatly according to multiplex signals (\eg retailer, address, payment time) and the high-order interaction of these signals.

Existing works in the transportation domain~\cite{li2018multi, hu2020stochastic, hong2020heteta} either assume all origins and destinations can be observed for spatial network construction or make straightforward use of the raw features.
Graph-based methods~\cite{li2018multi, hong2020heteta} exploit the power of graph neural networks (GNNs) to learn the representations of trips. However, these approaches require that all addresses in the OD graph are present during model training, making them hard to be generalized to unseen trips.
Another line of work~\cite{wu2019deepeta, wang2019asimple} utilizes deep neural networks to predict ETA based on the raw features. However, they failed in capturing the high-order semantic information that occurred in the interactions between the attributes of a package (\eg retailer, addresses, and payment time), which may also affect the accuracy of ETA prediction. The closest study is the food delivery time estimation conducted by Alibaba group on Ele.me~\cite{zhu2020order}. However, it focuses on city-level food delivery and highlights its contributions on feature engineering of existing models.

In this paper, we propose an inductive graph transformer (IGT) for the estimation of package delivery time on a national level. 
Contrary to currently coupled graph transformers~\cite{hu2020heterogeneous, dwivedi2020generalization} that leverage transformer as a substitution of information aggregator in GNN, we decouple the GNN and transformer pipeline in IGT. 
Specifically, we use transformer as a regression function and frame collected orders into sentences of NLP.
Analogous to word in a sentence for language processing, we denote retailer, addresses, and payment time as ``element'' in an order.
In this way, we can represent orders as a heterogeneous graph with different types of nodes (\ie element) and capture the high-order semantic relations with information propagation of GNN.
Furthermore, we adapt our model to large-scale scenarios by simplifying GNN with the removal of non-linear activation and linear transformation matrix.
In the following sections, we will be using the terms ``element'' and ``node'' interchangeably based on the discussed context.
With our simplified GNN, the embeddings of unseen elements in an order can be approximated from the observed neighbor elements.
In practice, orders have periodicity and recurring natures. 
We capture the temporal evolution of elements with a gated recurrent unit (GRU)~\cite{cho2014learning}.
Finally, we fuse the node embeddings and raw features of orders into a transformer-like architecture for ETA prediction.

In summary, the contributions made in this work are as follows:
\begin{itemize}
	\item We propose an inductive graph transformer for ETA prediction in package delivery. Instead of coupling transformer as an alternative attention mechanism for neighborhood information aggregation in GNNs, we use a decoupled architecture that feeds both extracted heterogeneous information from GNNs and raw features of orders into transformers for ETA prediction. 
	\item We design a simplified heterogeneous graph convolutional network that can encode high-order semantic information of orders into latent element embeddings and dynamically update element embeddings with GRU. Specifically, we remove the non-linear activation and the linear transformation matrix of the current graph convolutional network to adapt to large-scale industrial scenarios. As a result, only the linear operator for information aggregation and propagation is retained in IGT.
	\item We conduct comprehensive experiments on two large-scale logistics datasets, and the experimental results show that the proposed IGT model is significantly superior to the existing methods.
\end{itemize}

It is worth noting that although we mainly focus on the ETA prediction for package delivery, the proposed IGT model could be applied to any other similar scenarios.

\section{Related Work}

\subsection{Estimated Time of Arrival}

Previous ETA prediction methods can be categorized into two main groups: 1) route-based methods and 2) OD-based models. 

\subsubsection{Route-based Methods}
Apparently, route-based models take route information into consideration.
With the available of route information, a line of work uses (graph) neural networks to encode the spatial and temporal dependency.   
For example, WDR~\cite{wang2018learning} devises a Wide-Deep-Recurrent (WDR) architecture based on sequential features to formulate ETA as a regression problem. 
CompactETA~\cite{fu2020compacteta} and ConSTGAT~\cite{fang2020constgat} exploit graph attention networks (GAT)~\cite{velivckovic2018graph} on a spatio-temporal graph for representation learning.
HetETA~\cite{hong2020heteta} models different turning directions into a heterogeneous information graph and then learns the representation of the graph for ETA prediction. 
Note that the road structure in HetETA is assumed to be static, which is different from the package delivery in which graphs evolve dynamically. 
In logistics system, DeepETA uses a spatial-temporal sequential neural network to capture the regularity of the delivery pattern and the sequence of packages in a delivery route~\cite{wu2019deepeta}.
The above networks assume the nodes are present in the training of the embeddings and do not apply to inductive settings.

\subsubsection{OD-based Methods}
Due to privacy consideration, it is usually difficult to obtain the path information in many application scenarios. 
Under such a setting, OD-based methods that utilize the features of origin and destination are proposed for ETA prediction.
In~\cite{wang2019asimple}, the authors propose a simple model to predict travel time of a given trip by weighing the travel time of neighbors. The neighbors are filtered according to both the spatial information of origin and destination and the temporal information of trip starting time. 
Li \textit{et al.} develop a novel representation learning framework for travel time prediction, which uses road network, spatio-temporal smoothing prior, and additional trip attributes to represent the real world trip attributes~\cite{li2018multi}. 
The model can distill road network information in historical trips for accurate ETA prediction. However, it assumes the path information can be observed in historical trips, which is lack of generalization.
The most relevant existing work is the ETA prediction for food delivery~\cite{zhu2020order}, which is as complex as our problem because it is dependent on numerous features from couriers, restaurants, and traffic. However, this work mainly focuses on the exploitation of feature engineering to feed existing models.

\subsection{Graph Transformers}
Transformer~\cite{vaswani2017attention}, GNNs, and their variants have achieved great success in many fields.
A line of work has been proposed recently to fuse the two as a graph transformer. For example, Yun \textit{et al.} develop graph transformer networks on heterogeneous graphs targeting to transform a given heterogeneous graph into a meta-path based graph and then perform convolution consequently~\cite{yun2019graph}. 
Notably, their focus behind the use of attention framework is for interpreting the generated meta-paths. 
Another line of transformer on heterogeneous information networks is developed in~\cite{hu2020heterogeneous}. 
The proposed graph transformer uses transformer to substitute the attention mechanism in GAT.
Furthermore, Zhou \textit{et al.} propose a transformer based generative model which generates temporal graphs by directly learning from dynamic information in networks~\cite{zhou2020data}.
In~\cite{dwivedi2020generalization}, the authors propose a generalization of transformer neural network architecture for homogeneous graphs of arbitrary structure.
These existing methods generally employ the attention mechanism in transformer for information aggregation.
It is difficult to apply them to inductive graphs.

\section{Problem Statement}
This paper targets to solve the ETA prediction problem based on historical order delivery information. 
In our scenario, the platform will generate an order either after consumers have paid for their products or when the consumers are viewing a product page.
We formally define an order and the problem of delivery time estimation as follows.

\textit{Definition 3.1: Order}.
An order is represented as a tuple: $\mathbf{x}_i=(r_i, o_i, d_i, t_i)$, where $r_i$ denotes the retailer, $o_i, d_i$ are the origin location and destination of the order, respectively.
$t_i$ is the payment time of this order.
Each element in an order tuple is coupled with a hand-crafted dynamic feature vector describing its statistical information.
Thus, we can denote the dataset with $N$ historical orders by $\mathbf{X} = \{\mathbf{x}_i | i = 1, 2, \cdots, N\}$, and the associated delivery time by $\mathbf{y} = \{y_i | i = 1, 2, \cdots, N\}$.
A delivery time $y_i$ usually is calculated from the payment to the time when the order is signed. 

\textit{Definition 3.2: Problem of Delivery Time Estimation}.
Given a set of historical orders $\mathbf{X}$, the delivery time $\mathbf{y}$ and the input features $\mathbf{Z}$, we aim to estimate the delivery time $y_q$ of a query order $\mathbf{x}_q$.

\begin{figure*}[htp]
	\centering
	\includegraphics[trim=20 20 20 30, clip, width=0.9\textwidth]{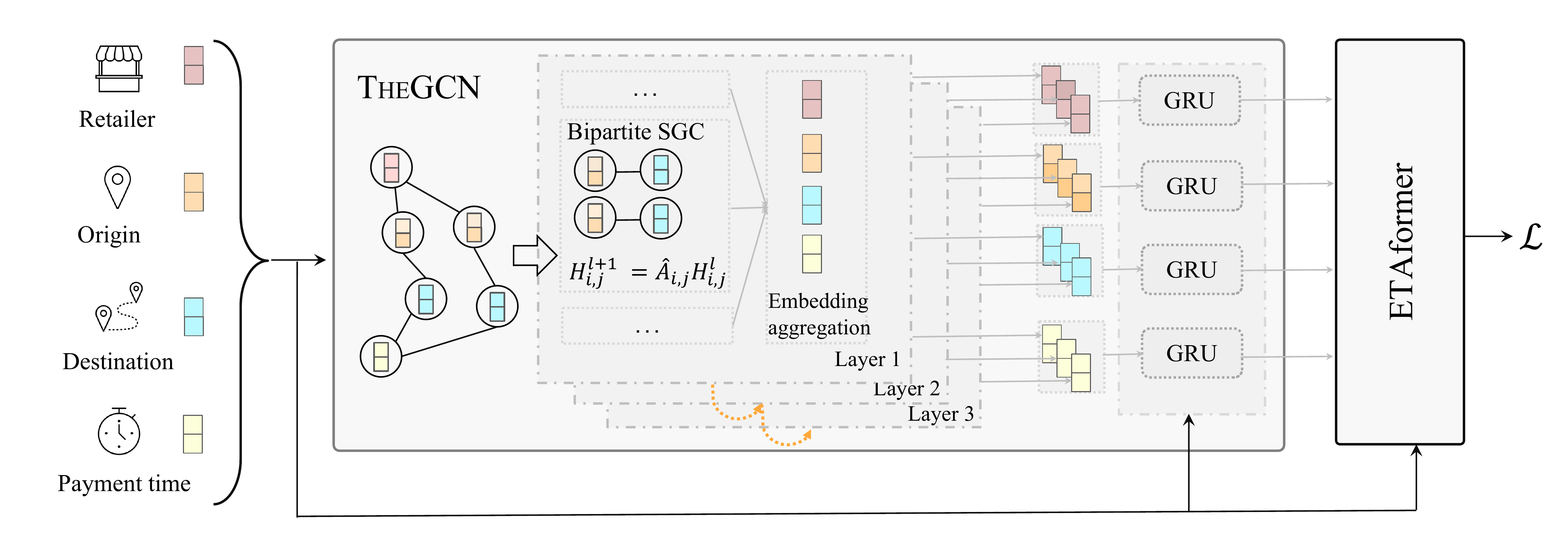}
	\caption{The overall framework of the proposed IGT model. There are two main modules in IGT: 1) T{\scriptsize HE}GCN (\ie Temporal and Heterogeneous GCN) and 2) ETAformer (\ie ETA Prediction with Transformer).}
	\label{fig:igt}
\end{figure*}

\section{Inductive Graph Transformer}
Since graph convolutional network (GCN)~\cite{kipf2017semi} was first outlined to generalize convolutional neural networks (CNNs) on graph structured data, GCN and its variants have been successfully applied to various domains~\cite{wang2019fdgars, yao2019graph, cheng2020skeleton, zhou2021selfcf, zhou2022bootstrap, zhou2022layer}.
As the order graph can be easily constructed from the historical orders. 
Each element in an order tuple is represented as node in the order graph, and two nodes are linked if they have occurred in the same order.
Considering the computational complexity of dealing with large-scale order graphs, we further limit the links to adjacent elements of an order. 

\textit{Definition 4.1: Construction of Heterogeneous Graph}.
Specifically, given an order $(r_i, o_i, d_i, t_i)$, the retailer node $r_i$ can only connect to the sender address node $o_i$. The sender address node $o_i$ can link to both the retailer node $r_i$ and the receiver address $d_i$, and so forth. That is, each order will generate only 6 edges instead of 12.
In this way, the dense of the graph can be cut down by half. Meanwhile, the information from one node can be propagated to any other nodes via stacking more convolutional layers of GCN. 

With the constructed heterogeneous graph, we can decouple the graph into a set of bipartite graphs and perform graph convolution on the bipartite graphs. 
The transformation operator and activation layer in GCN pose prohibitive cost for large-scale graphs. We simplify the graph convolutional layer with linear computation. 
However, apart from the structural information, the nodes in the graph may exhibit different temporal patterns. 
We adopt GRU to analyze temporal correlations on time-series axis at the node-level.
GRU solves the vanishing gradient problem of a standard recurrent neural network (RNN) by using update gate and reset gate.
Compared with the long short-term memory (LSTM) network, GRU is faster to train due to the fewer number of weights and parameters to update during training.
The output of GRU as well as the raw input features are feed into a  transformer variant for final prediction.
Figure~\ref{fig:igt} shows the overall framework of the proposed IGT model. It consists of two main modules: 1) T{\scriptsize HE}GCN (\ie \underline{T}emporal and \underline{He}terogeneous \underline{GCN}) and 2) ETAformer (\ie \underline{ETA} Prediction with Trans\underline{former}s). Next, we introduce the details of each component.

\subsection{T{\small HE}GCN}

Suppose $\mathcal{G} = (\mathcal{V},\mathcal{E})$ be the order graph derived from historical orders, where $\C{V}$ and $\C{E}$ denote the node set and edge set respectively.
Each node $v \in \C{V}$ and each edge $e \in \C{E}$ are associated with type mapping functions.
Our scenario includes four types of nodes and one type of edges. 
Hence, the constructed graph will be a heterogeneous graph.
We first introduce how GCN works on homogeneous graphs, then we show how to apply the convolutional operator to heterogeneous graphs.
A typical GCN on homogeneous graphs that recursively updates the hidden embedding $\mathbf{H}^{l}$ at the $l$-th layer is defined as:
\begin{equation}
\mathbf{H}^{l} = \sigma\left( \hat{\mathbf{A}}\mathbf{H}^{l-1}\mathbf{W}^{l} \right),
\label{eq:gcn_va}
\end{equation}
where $\mathbf{W}$ is the learnable transformation matrix and $\sigma(\cdot)$ is a non-linear function, \eg the ReLu function. $\hat{\mathbf{A}} = \hat{\mathbf{D}}^{-1/2}(\mathbf{A}+\mathbf{I})\hat{\mathbf{D}}^{-1/2}$ is the re-normalization of the adjacency matrix $\mathbf{A}$.
$\hat{\mathbf{D}}=\mathrm{diag}(d_1, \cdots, d_{|\mathcal{V}|})$ is the diagonal degree matrix of $\mathbf{A}+\mathbf{I}$, where each entry on the diagonal is equal to the row-sum of the adjacency matrix $d_i = 1+\sum_j a_{ij}$.
Each node $v_i$ in the graph has a corresponding $D$-dimensional latent feature vector $\mathbf{h}_i \in \mathbb{R}^D$. The initial feature matrix $\mathbf{H}^{0} = [\mathbf{h}_0,\cdots, \mathbf{h}_{|\mathcal{V}|}]^T$ stacks $|\mathcal{V}|$ feature vectors on top of one another.

In order to apply the graph convolution operator on heterogeneous graphs, a couples of GCN variants are proposed in other domains~\cite{zhu2020hgcn, ragesh2021hetegcn}. However, they are both computationally expensive and prone to over-fitting.
Inspired by the study of simplified homogeneous SGC~\cite{wu2019simplifying}, we design a heterogeneous GCN which towards computationally efficiency without compromising the performance.
Compared with SGC~\cite{wu2019simplifying}, we further remove its transformation matrix to reduce the parameter search space. This allows our model can be trained with large-scale datasets.
Given the heterogeneous graph $\mathcal{G}$, we first construct a set of bipartite subgraphs based on the combinations of the node types. 
Suppose the number of node types is $n$, the number of bipartite subgraphs will not exceed $\binom{n}{2} = \frac{(n)!}{2! (n-2)!}$.
In detail, we construct a sub-graph $\mathcal{G}_{i,j}$ from the original graph $\mathcal{G}$ based on the two node types $i$ and $j$. In $\mathcal{G}_{i,j}$, the type of a node belongs to either $i$ or $j$, and any two nodes in subgraph $\mathcal{G}_{i,j}$ will be linked if they are connected in the original graph $\mathcal{G}$. As $\mathcal{G}_{i,j}$ is undirected, we have $\mathcal{G}_{i,j} = \mathcal{G}_{j,i}$.
We denote the adjacency matrix of $\mathcal{G}_{i,j}$ as $\mathbf{A}_{i,j}$:
\begin{equation}
\mathbf{A}_{i,j} =
\begin{pmatrix}
\mathbf{0} & \mathbf{R} \\
\mathbf{R^T} & \mathbf{0}
\end{pmatrix},
\end{equation}
and each entry $R_{uv} \in \mathbf{R}$ is 1 if node $u \in \mathcal{G}_{i,j}$ is connected with node $v \in \mathcal{G}_{i,j}$ in $\mathcal{G}$; Otherwise, $R_{uv}$ is set to 0. 
Following~\cite{kipf2017semi}, we also perform the re-normalization trick on $\mathbf{A}_{i,j}$, resulting as $\hat{\mathbf{A}}_{i,j}$.
To perform convolution operation on bipartite graphs, we update the vanilla GCN in Eq.~\eqref{eq:gcn_va} as:
\begin{equation}
\mathbf{H}_{i,j}^{l} = \hat{\mathbf{A}}_{i,j} \mathbf{H}_{i,j}^{l-1},
\label{eq:gcn_big}
\end{equation}
where $\mathbf{H}_{i,j}^{l}$ denotes the embeddings of nodes in $G_{i,j}$ at the $l$-th GCN layer.
Suppose the number of nodes with type $i$ is $N_i$, then $\mathbf{H}_i \in \mathbb{R}^{N_i \times D}$ is the feature matrix of node type $i$ by stacking this type of nodes on top of one another.
Analogously, $\mathbf{H}_{i,j}^{l} \in \mathbb{R}^{(N_i+N_j) \times D}$.
Note that Eq.~\eqref{eq:gcn_big} simplifies the vanilla GCN by removing both the collapsing weight matrix and the non-linear layer, which further simplifies SGC~\cite{wu2019simplifying} with the removal of the non-linear activation.
Meanwhile, the reduction of trainable parameters prevents the model from over-fitting. The linear information propagation enables the model to be computationally efficient. We name this simplified graph convolution for bipartite graphs as Bipartite SGC in Figure~\ref{fig:igt}.

As one type of nodes may be involved in multiple bipartite graphs and participate the information propagation simultaneously, we aggregate the propagated information for each type of nodes within each layer as follows,
\begin{equation}
\mathbf{H}^l_{i} =  \mathrm{AGGREGATE}(\{\mathbf{H}_{i,j}^l[:N_i] \big| ~j = 1, ..., n, \mathrm{~and~} \mathcal{G}_{i,j} \neq \emptyset\}),
\label{eq:inlayer_agg}
\end{equation}
where $\textrm{AGGREGATE}$ is an aggregation function, and $n$ is the number of node types.
Here, we use the sum aggregation.
Other aggregation operators~\cite{corso2020principal} can also be applied for aggregating the embeddings of nodes belonging to the same type. In Eq.~\eqref{eq:inlayer_agg}, we use the slice operator to extract the embeddings of $i$-type nodes.

One benefit of such information aggregation within each layer is that the information from different types of nodes can be completely integrated with each other. 
The integrated information is then fed into the next layer for propagation.
Another benefit is that T{\scriptsize HE}GCN updates the embeddings of all nodes instead of a batch of nodes in each iteration to avoid the memory staleness problem~\cite{kazemi2020representation}.

With the information obtained in each layer, we update the nodes' final embeddings at time $t-1$ by aggregating all intermediate embeddings. That is:
\begin{equation}
\mathbf{H}_{i}^{t-1} =  \mathrm{AGGREGATE}(\mathbf{H}_{i}^0, \mathbf{H}_{i}^1, ..., \mathbf{H}_{i}^L),
\label{eq:layer_agg}
\end{equation}
where $L$ is number of propagation layers.
The aggregation function $\textrm{AGGREGATE}$ can be any functions discussed in~\cite{corso2020principal}.
Here, we use the mean aggregation for final node updating.

To further fuse the temporal information into the learnt embeddings $\mathbf{H}^{t-1}_{i}$ at time $t-1$, we use the gating mechanisms to control and manage the flow of information. 
\begin{equation}
\mathbf{H}^{t}_{i} = \textrm{UPDATE}(\mathbf{H}^{t-1}_{i}, \mathbf{Z}^t_{i}), 
\label{eq:gru_update}
\end{equation}
where $\mathbf{Z}^t_{i}$ is the raw input feature vectors of $i$-type elements in orders at time $t$.
Here, $\textrm{UPDATE}$ is a learnable update function, \eg a recurrent neural network such as LSTM or GRU.
In this work, we adopt the GRU network because it owns fewer number of gates and is as effective as LSTM.

Note that we use a normalized adjacency matrix constructed from historical orders to avoid data leakage in model training stage. However, in inference stage, we use the normalized adjacency matrix built upon both training and test data~\cite{hamilton2017inductive}.
Under inductive setting, our proposed T{\scriptsize HE}GCN can fuse information from its neighbors within the same order $\mathbf{x}_i$ to learn the embeddings of unseen nodes. In extremely case, all nodes in $\mathbf{x}_i$ are never observed in training, which means that no neighbor information can be used. T{\scriptsize HE}GCN adopts GRU to update the node's embeddings from raw input features. Hence, raw features will be used to estimate the delivery time of this order.

\begin{algorithm}
	\caption{Pseudo-code for Inductive Graph Transformer (IGT)} \label{alg:alg}
	\begin{algorithmic}[1]
		\State \textbf{Inputs:} Set of orders $\mathbf{X}$, the associated delivery time $\mathbf{y}$ and input feature vectors $\mathbf{Z}$.
		\State Construct heterogeneous graph $\C{G}$ from $\mathbf{X}$.
		\State Initialize latent embeddings $\B{H}^0$ of nodes in $\C{G}$ with Xavier.
		\State Extract bipartite subgraphs with node type $i$ and $j$ $\C{G}_{i,j}$ from $\C{G}$ and the latent embeddings $\B{H}_{i,j}^0$ from $\B{H}^0$.
		\For{$\mathbf{X}^t ~ in ~ \mathbf{X}$}\Comment{load a batch} 
		\For{$l$ in [1, ..., $L$]} \Comment{layer-wise propagation}
		\For{$i$ in [1, ..., $n$]}
		\For{$j$ in [1, ..., $n$]}
		\State $\mathbf{H}_{i,j}^{l} = \hat{\mathbf{A}}_{i,j} \mathbf{H}_{i,j}^{l-1}$ \Comment{bipartite SGC}
		\EndFor
		\EndFor
		\For {$i$ in [1, ..., $n$]}
		\State $\mathbf{H}^l_{i} =  \mathrm{AGGREGATE}(\{\mathbf{H}_{i,j}^l[:N_i]\})$ \Comment{Eq.~\ref{eq:inlayer_agg}}
		\EndFor
		\EndFor \Comment{end layer propagation loop}
		\For{$i$ in [1, ..., $n$]}
		\State $\mathbf{H}_{i}^{t-1} =  \mathrm{AGGREGATE}(\mathbf{H}_{i}^0, \mathbf{H}_{i}^1, ..., \mathbf{H}_{i}^L)$ \Comment{Eq.~\ref{eq:layer_agg}}
		\State $\mathbf{H}^{t}_{i} = \textrm{UPDATE}(\mathbf{H}^{t-1}_{i}, \mathbf{Z}^t_{i})$ \Comment{Eq.~\ref{eq:gru_update} for temporal update}
		\EndFor
		\State $\hat{\mathbf{y}}^t = \mathrm{ETAformer}(\mathbf{H}^{t}, \mathbf{Z}^t)$ \Comment{ETA prediction} 
		\State $\mathcal{L} = \mathrm{MAE}(\mathbf{y}^t, \hat{\mathbf{y}}^t)$ \Comment{loss}
		\State $\mathcal{L}$.backward() \Comment{back-propagate}
		\EndFor
	\end{algorithmic} 
\end{algorithm}

\subsection{ETAformer}
The standard transformer is designed for word prediction, and a sentence is treated as a fully connected graph.
An order in our scenario is analogy to a sentence in natural language processing (NLP) with less than tens or hundreds of elements. 
As such, the ETA prediction is substantially amenable to transformers.
Large transformer models can be trained on such fully connected graphs.
Furthermore, each word attends to each other word in a sentence in the transformer architecture, which enables the high-order interactions among words. 
In NLP, the prediction of a word in a sentence can vary with context, other words etc.
Equivalently, in our scenario, the ETA prediction relies not only on the raw features from retailers, addresses, payment time, but also the high-order interactions of these features. 
However, the standard transformer assume words are encoded within the same embedding space. 
In practice, the nodes in heterogeneous graph may vary in feature size as they belonging to different types.
Thus, to encapsulate the input feature vectors into transformer, we first need to reshape the feature space. 
That is:
\begin{equation}
\mathbf{e}_i = f(\mathbf{z}_i, \mathbf{h}^t_i),
\label{eq:emb_map}
\end{equation}
where $\mathbf{z}_i$ is the raw feature of node $i$, $\mathbf{h}^t_i$ is the output embedding of node $i$ from T{\scriptsize HE}GCN, and $f$ is a mapping function to align nodes with the same dimensionality.
An intuitive function is using multilayer perceptron (MLP) to map features of different dimensional spaces to the same space. 
However, it introduces additional parameters to the model, so we adopt zero-padding on each node of an order.
Finally, we reshape the order into $\mathbf{x}^i \in \mathbb{R} ^{n \times W}$, where $n$ is the number of elements in an order and $W$ is the padded embedding size of each element.

\begin{figure}[ptb]
	\centering
	\includegraphics[angle=90, trim=20 20 80 20, clip, width=0.8\linewidth]{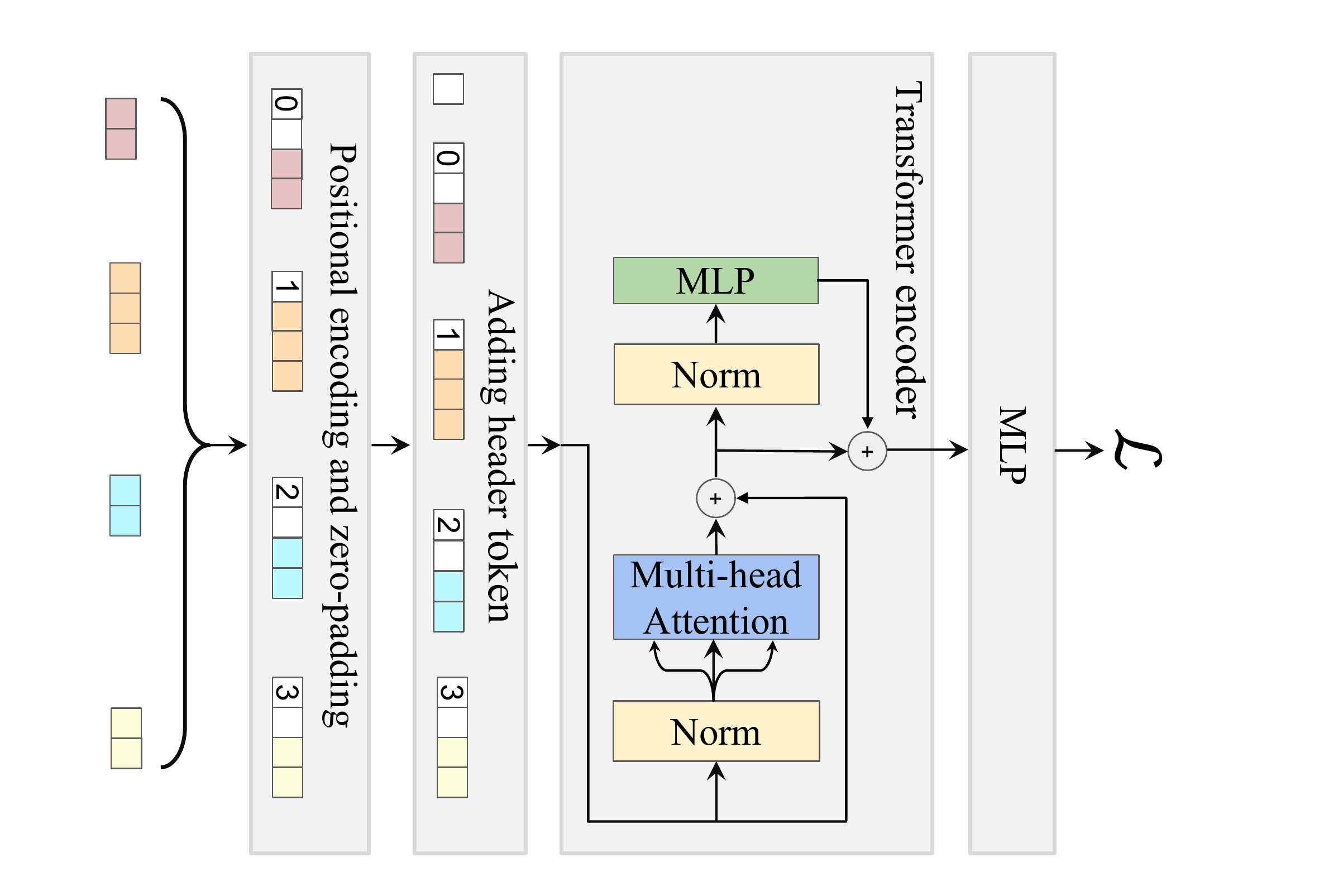}
	\vspace{-10pt}
	\caption{The architecture of ETAformer.}
	\label{fig:etaformer}
	\vspace{-10pt}
\end{figure}

For capturing the order-level information, we prepend a learnable header token similar to that of BERT~\cite{devlin2019bert} and ViT~\cite{dosovitskiy2021image}.
In implementation, we use a MLP with one hidden layer for token learning. On positional encodings, existing graph transformers design various methods to preserve the distance information between nodes. Our decoupled architecture utilizes the standard positional encoding of transformer. Because an order $\mathbf{x}_q$ in our scenario can be naturally framed as a sequence of elements. The header token $\mathbf{e}_q^{header}$, positional encodings $\mathbf{e}_q^{pos}$, and mapped node embeddings $\mathbf{e}_q$ of $\mathbf{x}_q$ are then fed into the transformer variants.
Upon the output, we finally use a layer normalization (LN) and a single layer MLP to predict the ETA, as shown in Figure~\ref{fig:etaformer}.

Formally, we denote the raw feature of a query order $\mathbf{x}_q$ as $\mathbf{z}_q = (z_q^1, z_q^2, ..., z_q^n)$, where $z_q^i$ is the raw feature of $i$-th element in the order.
The ETA prediction of order $\mathbf{x}_q$ based on the transformer-like models can be presented as follows,
\begin{equation}
\begin{aligned} 
& \mathbf{e}_q = f(\mathbf{z}_q, \mathbf{h}_q^t), \\
& \hat{\mathbf{e}}_q^{header} = \mathrm{Transformer}(\mathbf{e}_q^{header}, \mathbf{e}_q^{pos}, \mathbf{e}_q), \\
& \hat{y}_q = \mathrm{MLP}\left( \mathrm{LN}(\hat{\mathbf{e}}_q^{header}) \right).
\end{aligned}
\end{equation}

\subsection{Model Training}
For training the proposed IGT model, we utilize the Adam~\cite{kingma2015adam} as the optimizer, and use the Mean Absolute Error (MAE) as our loss function. Considering a set of $N$ samples $\mathbf{y}=(y_1, y_2, \cdots, y_N)$ representing the ground truth and the associated predictions $\hat{\mathbf{y}}$, the MAE-based loss function is defined as follows:
\begin{equation}
\mathcal{L} = \mathrm{MAE}(\mathbf{y}, \hat{\mathbf{y}}) = \frac{1}{N}\sum^N_{i=1}|y_i - \hat{y}_i|.
\label{eq:mae}
\end{equation}
The details of the optimization algorithm used to learn the IGT model are summarized in Algorithm 1. 

\section{Experiments}

\begin{table}
	\caption{Statistics of experimental datasets.}
	\label{tab:data_statis}
	\def\arraystretch{0.9}
	\begin{tabular}{ l|c|c } 
		\hline
		Datasets & D1 & D2 \\
		\hline		
		\# of orders & 375,393 & 1,817,766\\
		\# of retailers & 4,729 & 8,741 \\
		\# of days & 110 & 47 \\
		validation/test days & (10, 15) & (5, 10)\\
		\hline
		\# of orders in test & 61,647 & 348,037 \\
		\# of orders with unseen retailers & 913 & 930 \\
		\# of orders with unseen addresses & 61 & 3 \\
		\hline
	\end{tabular}
\end{table}

\subsection{Experimental Datasets}
We utilize two large-scale logistics datasets collected from one of the world's largest e-commerce platforms to evaluate the performance of IGT as well as the baselines. 
\subsubsection{Data Description and Statistics}
The first dataset contains 0.38 million samples from 4,729 retailers (denoted by D1), ranging from January 3rd to April 22nd, 2021. 
The second dataset includes 1.8 million samples collected from 8,741 retailer (denoted by D2), ranging from June 3rd to July 22nd, 2021.
The two datasets are not only differentiate in graph size, collected time, but also in collection regions.
Both datasets are orders targeting a fixed receiving city.
Specifically, D1 is collected with all the orders sent to Weihai, China. D2 is for packages oriented to Hangzhou, China.
The population of the receiving city of Hangzhou is 3 times more than it is in Weihai.
From Table~\ref{tab:data_statis}, we can also observe that the retailers in D2 are relatively more active than those in D1 with an average of 208 orders.
For ETA prediction, we follow the chronological data splitting for training, validation, and test.
To be specific, we split the orders in the last 25 days of D1 by 10:15 as validation and test, the last 15 days of D2 by 5:10, orders from other days are used for training.
In Table~\ref{tab:data_statis}, we denote the unseen nodes (\eg retailers, addresses) as the nodes in test that never occurred in training.
Note that the datasets also collect the evolving features of retailer, payment time, sender address, and receiver address.

\begin{figure}[!h]
	\centering
	\begin{subfigure}[b]{0.235\textwidth}
		\includegraphics[trim=10 12 10 8, clip, width=\textwidth]{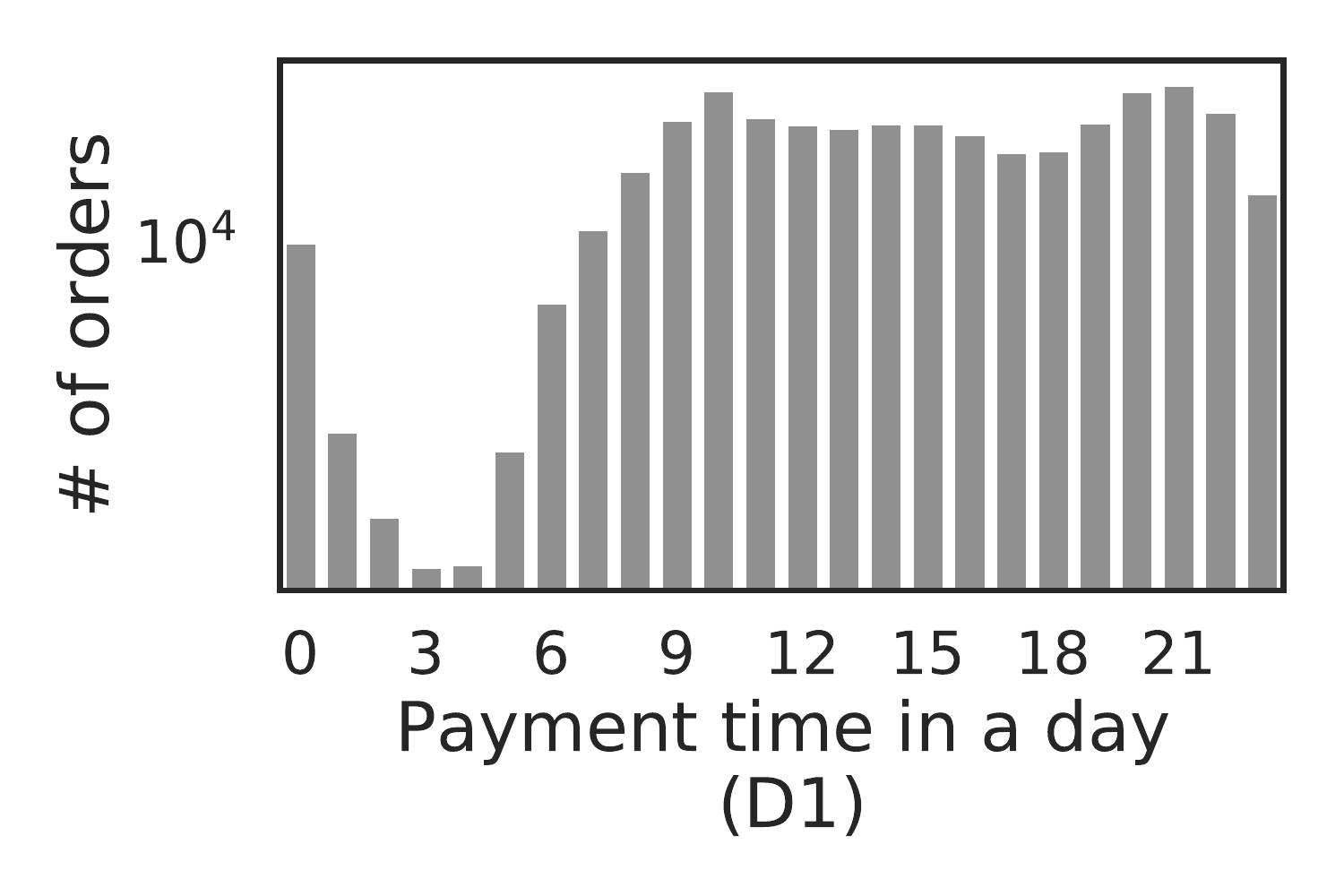}
	\end{subfigure}
	\hfill
	\begin{subfigure}[b]{0.235\textwidth}
		\includegraphics[trim=10 12 10 8, clip, width=\textwidth]{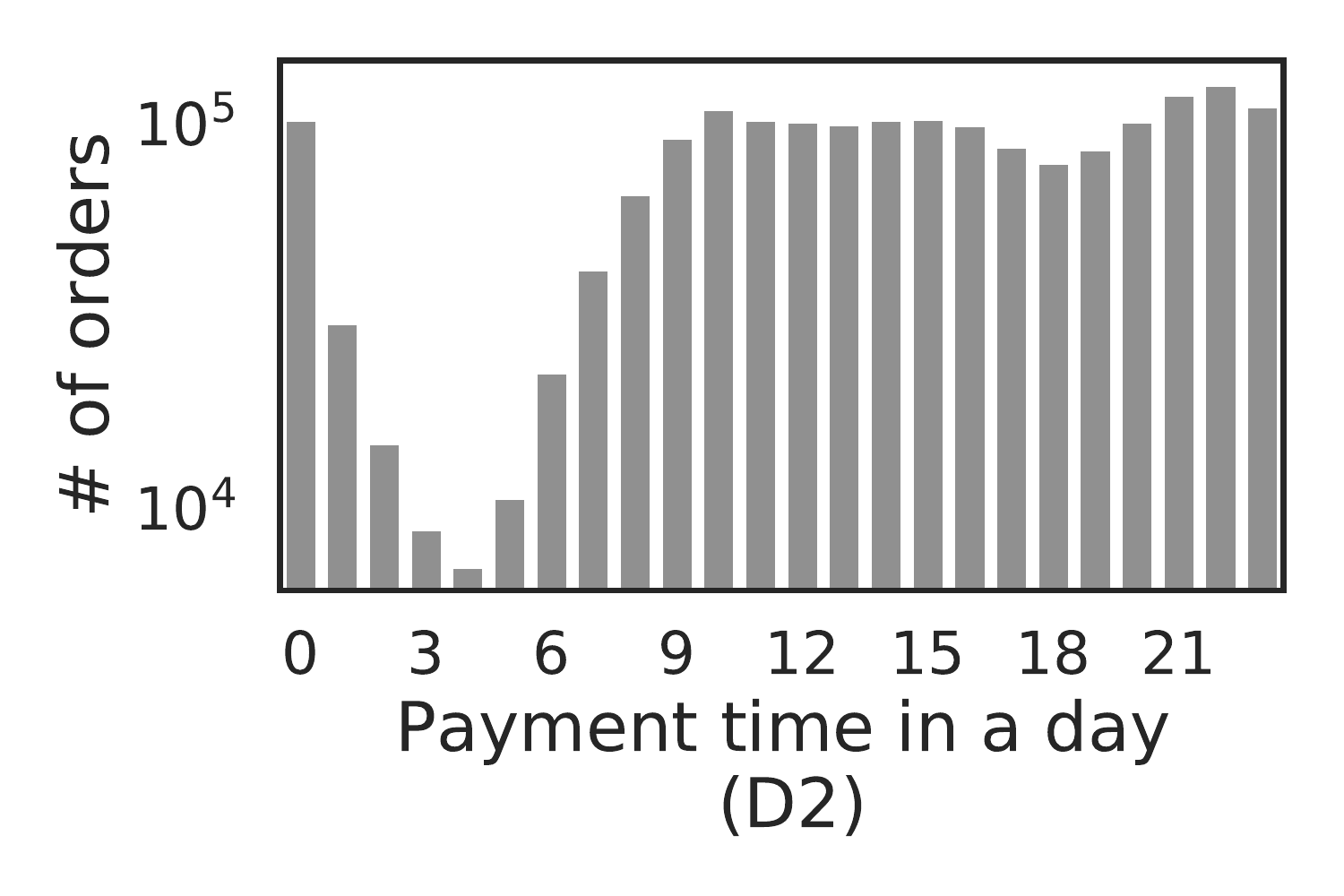}
	\end{subfigure}
	\caption{Diurnal pattern of orders. $y-$axis is in log scale.}
	\label{fig:time_dis}
	\vspace{-10pt}
\end{figure}

To facilitate our interpretation of the prediction results on different datasets, we plot the histogram of the diurnal pattern on order payment time in Figure~\ref{fig:time_dis}. 
From the plots, we can observe the payment time pattern is consistent with human regularity. 
The two datasets also vary in order volume, that is, the number of orders received in Hangzhou is almost an order of magnitude of Weihai at each payment time.

\subsection{Comparison Methods and Metrics}
We compare the proposed IGT model with the following baselines.
1). \textbf{Linear Regression (LR)} models the delivery time of orders by training a linear regression model based on the raw features of orders. We use the LR in scikit-learn~\cite{pedregosa2011scikit} for evaluation.
2). \textbf{XGBoost}~\cite{chen2016xgboost} is a gradient boosting decision tree based regression model. The input is the same as the LR model. We use the XGBRegressor in xgboost~\footnote{https://xgboost.readthedocs.io} with MAE as its evaluation metric.
3). \textbf{MURAT}~\cite{li2018multi} captures underlying road network structures as well as spatio-temporal prior knowledge for OD travel time estimation. The road information is not available in our dataset, the OD graph is built upon the sender and receiver address. We use both ResNet18 and deep feed forward neural networks (DNN) to map the learned representations to target tasks. In our scenario, we found DNN performs better than ResNet18. The reported results in our paper are based on DNN.
4). \textbf{xDeepFM}~\cite{lian2018xdeepfm} learns both low- and  high-order feature interactions in both explicit and implicit fashions in an extreme deep factorization machine framework. We transform the recommendation task into a regression task for ETA prediction.
5). \textbf{TEMP{\scriptsize \texttt{rel}}}~\cite{wang2019asimple} is an OD-based approach for travel time estimation. It averages the scaled travel times of all neighboring routes with similar origin and destination for ETA prediction.
6). \textbf{DeepETA}~\cite{wu2019deepeta} is an end-to-end network for last-mile package delivery estimation. It not only takes route features into consideration but also encodes the spatial and temporal information in deep neural networks for ETA prediction. As no route information is available in our data, we use the sender and receiver addresses as the route history. 
7). \textbf{HetETA}~\cite{hong2020heteta} targets to estimate the arrival time of taxis within a city. It uses gated convolutional neural networks and ChebNet~\cite{defferrard2016convolutional} to capture the relations between through spatial and temporal perspectives. This method relies on the trajectory information. Again, as our data lacks this type of information, we perform Het-ChebNets only on the road network. The road network is built up as the same as MURAT.

\subsection{Performance Comparison}
\begin{table*}
	\caption{Performance comparison on ETA prediction between our model and the baselines. We mark the best results on each dataset under each metric in \textbf{Bold} face, and the second best underlined. \textit{Improvement (\%)} is calculated between the best result achieved from the baselines and the result of our model.}
	\label{tab:perform}
	\def\arraystretch{0.9}
	\begin{tabular}{ l|l|c|c|c|c|c|c|c|c|r } 
		\hline
		Datasets & Metric & LR & XGBoost & MURAT & xDeepFM & TEMP{\scriptsize \texttt{rel}} & DeepETA & HetETA & IGT & \textit{Improvement (\%)}\\
		\hline
		\multirow{3}{*}{D1} & MAE(\textit{hours}) & 23.46 & 11.94 & 12.15 & \underline{11.85} & 17.19 & 16.61 & 13.41 & \textbf{10.53} & 11.14 \\
		& MAPE(\textit{\%}) & 31.28 & \underline{15.46} & 16.11 & 15.73 & 26.48 & 26.38 & 18.86 & \textbf{13.35} & 13.65 \\
		& MARE(\textit{\%}) & 34.02 & 17.32 & 17.62 & \underline{17.18} & 24.93 & 24.09 & 19.45 & \textbf{15.28} & 11.06 \\
		\hline
		\multirow{3}{*}{D2} & MAE(\textit{hours}) & 10.75 & 8.16 & 8.83 & \underline{7.99} & 10.53 & 14.58 & 10.57 & \textbf{7.39} & 6.13 \\
		& MAPE(\textit{\%}) & 26.21 & 17.24 & 19.31 & \underline{16.50} & 24.37 & 35.69 & 24.66 & \textbf{14.22} & 9.82 \\
		& MARE(\textit{\%}) & 22.92 & 17.39 & 18.83 & \underline{17.03} & 22.45 & 31.08 & 22.53 & \textbf{15.75} & 6.11 \\
		\hline
	\end{tabular}
\end{table*}

All models are trained on a single GeForce RTX 2080 Ti with 11 GB memory.
For traditional machine learning models, we feed full training data into LR, XGBoost, and TEMP{\scriptsize \texttt{rel}} for ETA prediction.
Other models are trained in batches with a size of 8192.
We monitor the loss over the validation set and apply early stopping if no improvement happens after 100 consecutive epochs. 
Considering the model convergence, the total epochs for training are fixed at 1000.
We initialize the latent embeddings $\mathbf{H}^0$ of nodes in $\mathcal{G}$ with the Xavier method~\cite{glorot2010understanding}, tune the number of layers $L$ of T{\scriptsize HE}GCN in $\{1, 2, 3, 4, 5\}$, and the embedding size $D$ of nodes in $\{16, 32, 64, 128, 256\}$.
We implement our model in PyTorch~\cite{paszke2019pytorch}.
Source code of IGT will be released upon acceptance.

Following~\cite{li2018multi}, we evaluate our model and all baselines under the metrics of MAE, Mean Absolute Percentage Error (MAPE), and Mean Absolute Relative Error (MARE). The definitions of MAPE and MARE are as follows, 
\begin{align}
\mathrm{MAPE}(\mathbf{y}, \hat{\mathbf{y}}) &= \frac{1}{N}\sum^N_{i=1}\big|\frac{y_i - \hat{y}_i}{y_i} \big|, \\
\mathrm{MARE}(\mathbf{y}, \hat{\mathbf{y}}) &= \frac{\sum^N_{i=1}|y_i - \hat{y}_i|}{\sum^N_{i=1} y_i}.
\end{align}
Analogous to MAE defined in Eq.~\eqref{eq:mae}, $\mathbf{y}$ denotes the ground truth of package delivery time, $\hat{\mathbf{y}}$ denotes the predicted delivery time, and $N$ denotes the number of samples in test data. 

Table~\ref{tab:perform} presents the performance comparison between the proposed IGT model and the baselines on two datasets. 
The results on three metrics show our proposed model is capable of improving the best baseline by more than 6\%.
Especially on dataset D1, IGT gains significant improvement over the best baseline by up to 11\%.
Compared with D2 in Table~\ref{tab:data_statis}, D1 is more sparse. 
The averaged number of orders for each retailer is 0.7 and 4.4 per day in D1 and D2, respectively.
The results between D1 and D2 indicate that IGT performs better with sparse graph.
However, sparsity may partially contribute to the performance. 
We analyze why the performance of the two datasets vary heavily through the perspective of information theory.

\begin{figure}[!h]
	\centering
	\includegraphics[trim=11 12 6 10, clip, width=0.5\textwidth]{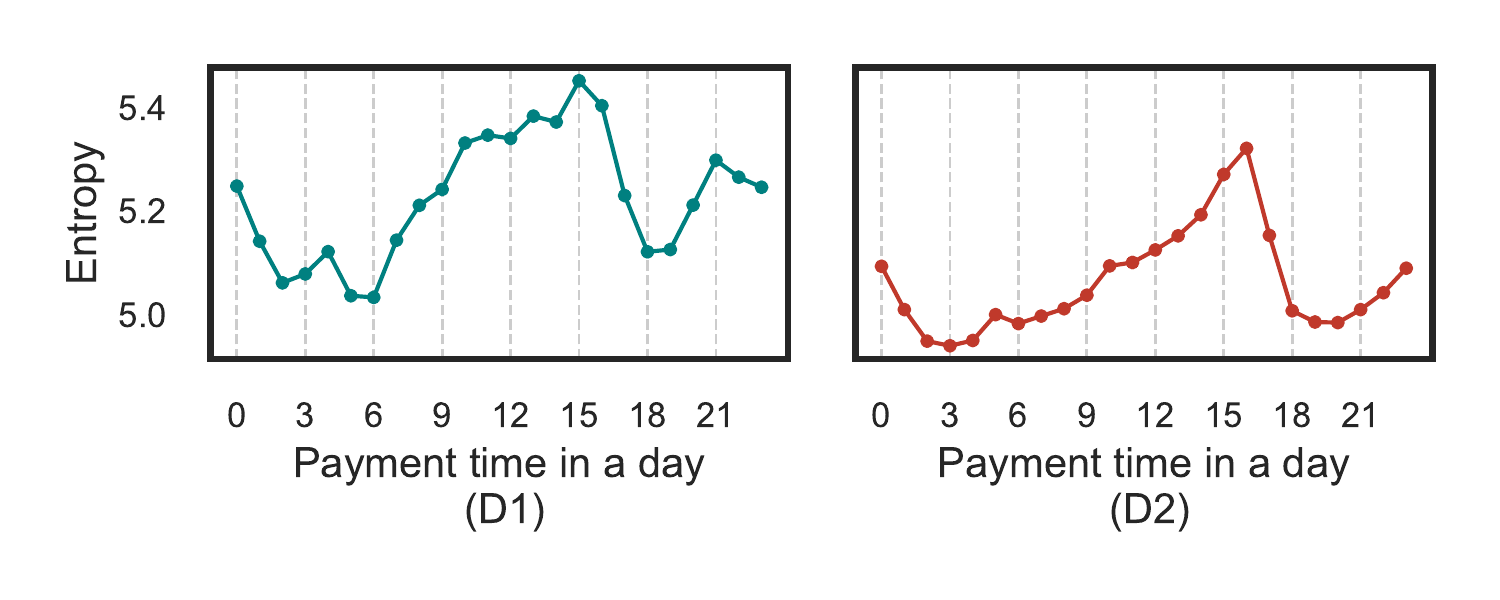}
	\caption{Entropy of the delivery time for orders placed at payment time.}
	\label{fig:entropy}
\end{figure}

We utilize entropy to quantify the entropy of the distribution of delivery time under a payment time.

\begin{equation}
p^t = -\sum_{y_i^t \in \mathbb{S}^t} p(y_i^t) \mathrm{log}p(y_i^t), 
\end{equation}
where $\mathbb{S}^t$ is the set of delivery time for orders placed at $t$.
The entropy is larger for orders with greater divergent delivery time that placed at the same time $t$.
As such, we plot the entropy of the two datasets with regard to the payment time in Figure~\ref{fig:entropy}.
The figure reveals dataset D1 has a larger averaged entropy value than that of dataset D2.
As a result, the MAE value for D1 is much larger than D2 for all models.
We will re-examine the variation of entropy in Section~\ref{sec:it_eta}.

As for the baselines, the performance of LR is poor mainly because it can only use the raw features for regression tasks. 
Other regression models such as XGBoost and xDeepFM can capture the interactions between features, thus achieving better performance.
TEMP{\scriptsize \texttt{rel}} improves LR on both datasets implying the importance of spatial (\ie longitude, latitude) and temporal information in ETA prediction.
The performance of MURAT, DeepETA and HetETA is compromising as they are designed to take advantage of route information for ETA prediction.
However, MURAT and HetETA utilize graph representation learning to learn the meaningful embeddings of OD links and achieve better performance over DeepETA.
Similarly, our model also uses graph learning (\ie GCN) to capture the structural information between orders.
Nevertheless, we focus on node representation learning, and a link that connects retailer, addresses and payment time is represented by concatenating the embeddings of these nodes.

\subsection{Ablation Study}
\label{sec:abs}
We perform ablative study on each component of IGT to examine its contribution in ETA prediction.
Recalling that IGT consists of two components, namely T{\scriptsize HE}GCN and ETAformer. 
The decoupled architecture of IGT enables us to perform ETA prediction on each component without the prerequisite of the other one.
First, to evaluate T{\scriptsize HE}GCN, we feed the concatenation of the latent embeddings obtained from T{\scriptsize HE}GCN into a single layer MLP for ETA prediction.
As for ETAformer, it can cope with the raw features directly. 
We report the performance on the two datasets under each metric in Table~\ref{tab:ab_study}.

\begin{table}[btp]
	\caption{Ablation study of IGT.}	
	\def\arraystretch{0.9}
	\label{tab:ab_study}
	\begin{tabular}{ l|l|c|c|c } 
		\hline
		Datasets & Metric & T{\scriptsize HE}GCN & ETAformer & IGT \\
		\hline
		\multirow{4}{*}{D1} & MAE(\textit{hours}) & 11.79 & 10.67 & 10.53  \\
		& MAPE(\textit{\%}) & 15.67 & 13.22 & 13.35 \\
		& MARE(\textit{\%}) & 17.09 & 15.48 & 15.28 \\
		& time/epoch(\textit{seconds}) & 2.66 & 10.47 & 13.34 \\
		\hline
		\multirow{4}{*}{D2} & MAE(\textit{hours}) & 7.99 & 7.66 & 7.39 \\
		& MAPE(\textit{\%}) & 16.62 & 15.43 & 14.22 \\
		& MARE(\textit{\%}) & 17.03 & 16.34 & 15.75 \\
		& time/epoch(\textit{seconds}) & 11.31 & 42.86 & 63.06\\
		\hline
	\end{tabular}
	\vspace{-10pt}
\end{table}

The results show that ETAformer achieve better performance than T{\scriptsize HE}GCN in ETA prediction.
Because ETAformer acts on inter-order, it further explores the high-order interactions between elements of an order.
Comparing the performance of T{\scriptsize HE}GCN with that of xDeepFM in Table~\ref{tab:perform}, we can observe that merely using of T{\scriptsize HE}GCN in IGT can obtain competitive performance to xDeepFM. 
Because T{\scriptsize HE}GCN is also able to extract the high-order information between orders by virtue of linear information propagation. 
The training time in Table~~\ref{tab:perform} demonstrates T{\scriptsize HE}GCN is significantly faster than ETAformer. 
The combination of T{\scriptsize HE}GCN and ETAformer gains significant improvements over the baselines.

\subsection{Hyper-parameter Study}
\begin{figure}[btp]
	\centering
	\begin{subfigure}[b]{0.235\textwidth}
		\includegraphics[trim=13 12 13 10, clip, width=\textwidth]{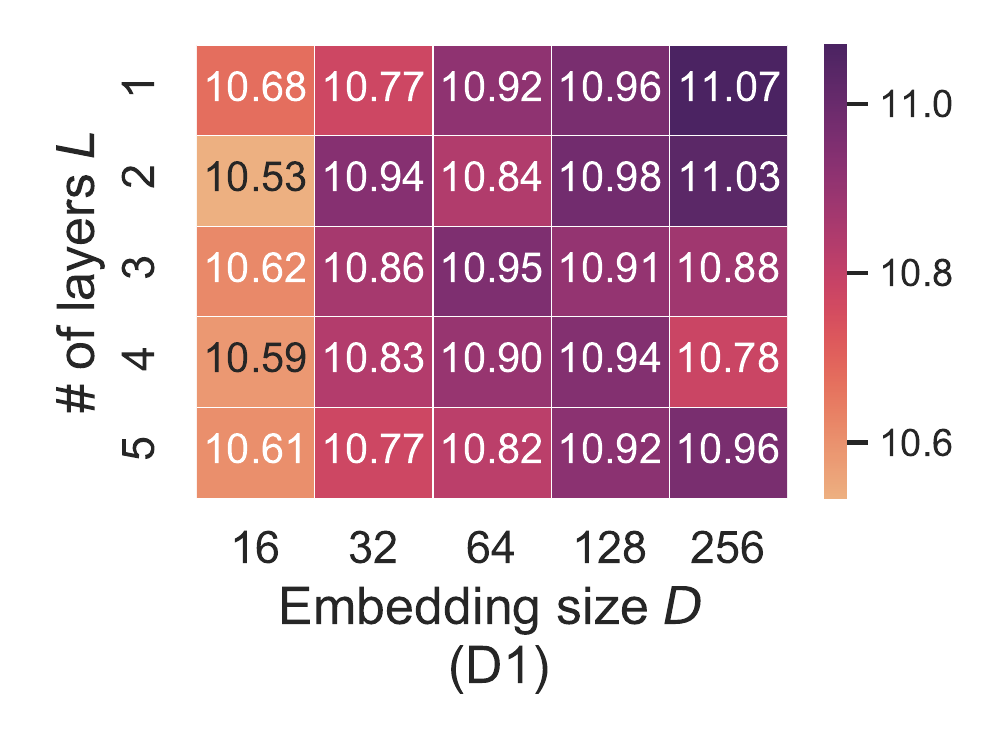}
	\end{subfigure}
	\hfill
	\begin{subfigure}[b]{0.235\textwidth}
		\includegraphics[trim=13 12 13 10, clip, width=\textwidth]{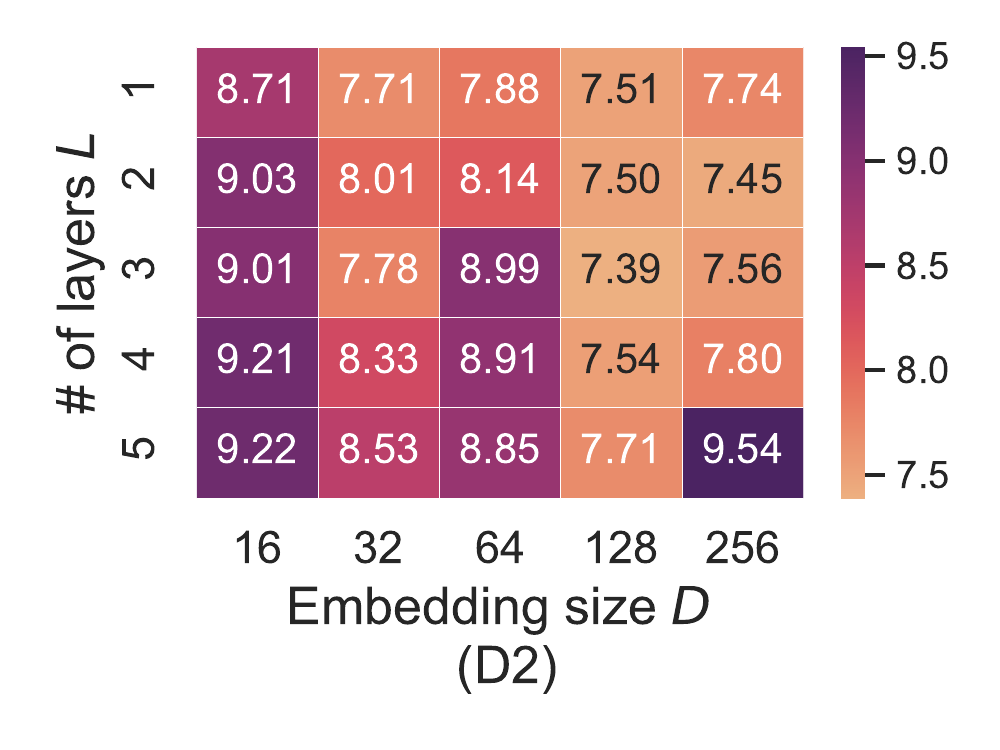}
	\end{subfigure}
	\caption{Hyper-parameter study of IGT.}
	\label{fig:hyper_heat}
\end{figure}

We examine how the number of layers $L$ and the embedding size $D$ of nodes in T{\scriptsize HE}GCN affect the performance of IGT.
We collect the MAE performance of IGT under the grid-search of hyper-parameters $L$ and $D$ on both datasets (D1 and D2) and plot the heatmap in Figure~\ref{fig:hyper_heat}.
We observe that the dimensionality $D$ of the hidden embeddings in T{\scriptsize HE}GCN has a more decisive impact on the ETA prediction.
Stacking too many layers $L$ in T{\scriptsize HE}GCN may result in over-smoothing~\cite{li2018deeper} and degenerate the prediction performance.
A good choice for the dimensionality of hidden embeddings may depend on the number of orders in the dataset. 

\subsection{Inductive and Transductive Learning of IGT}
\label{sec:it_eta}
We further analyze the prediction results with regard to the types of nodes and the number of orders in each node type.
Especially, we examine the prediction performance of IGT under inductive setting where nodes have no historical orders.
In our analysis, we compare IGT with xDeepFM to elucidate the use cases where IGT may provide a significant edge.

\begin{figure}[bpt]
	\centering
	\begin{subfigure}[b]{0.235\textwidth}
		\includegraphics[trim=15 22 15 18, clip, width=\textwidth]{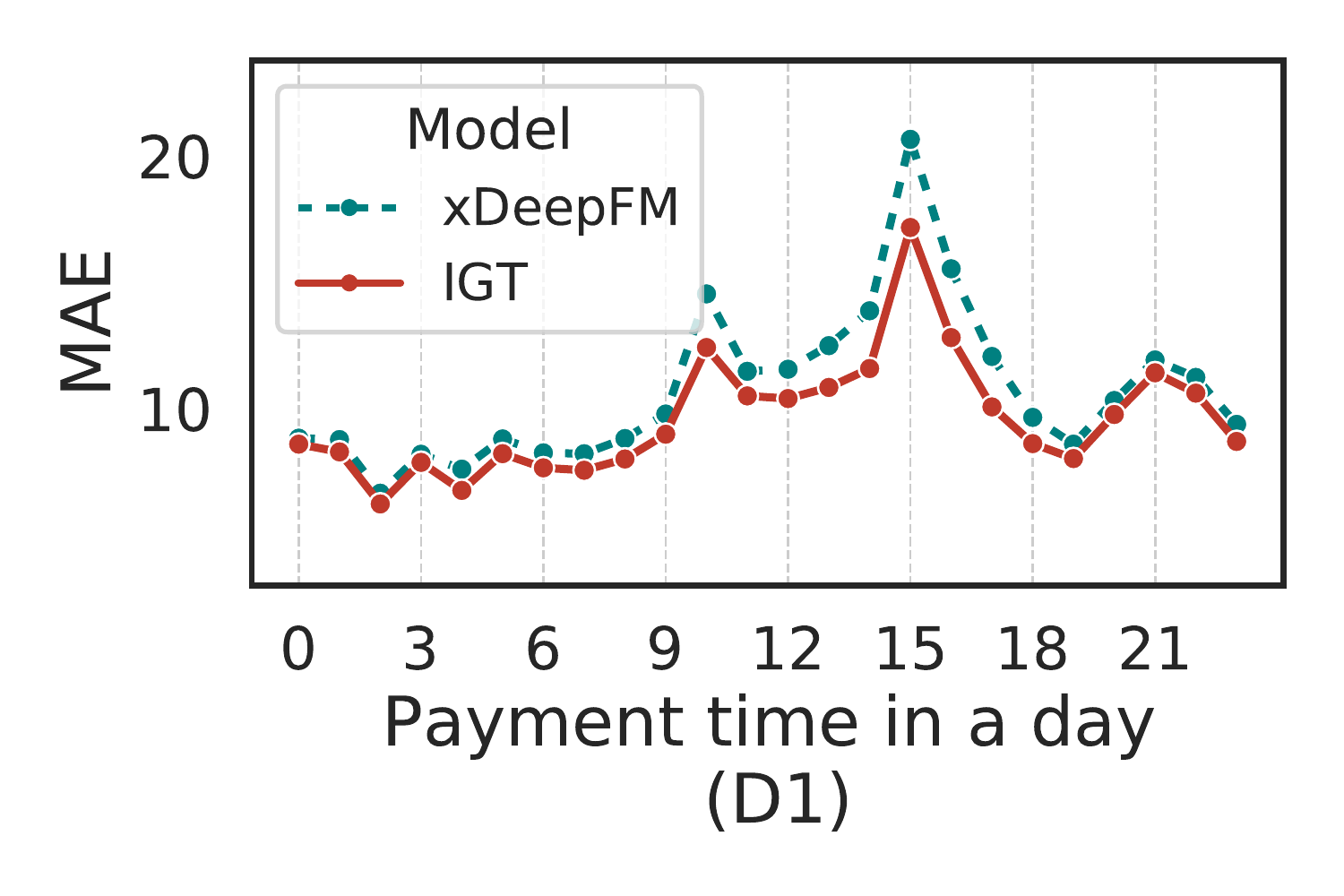}
	\end{subfigure}
	\hfill
	\begin{subfigure}[b]{0.235\textwidth}
		\includegraphics[trim=15 22 15 18, clip, width=\textwidth]{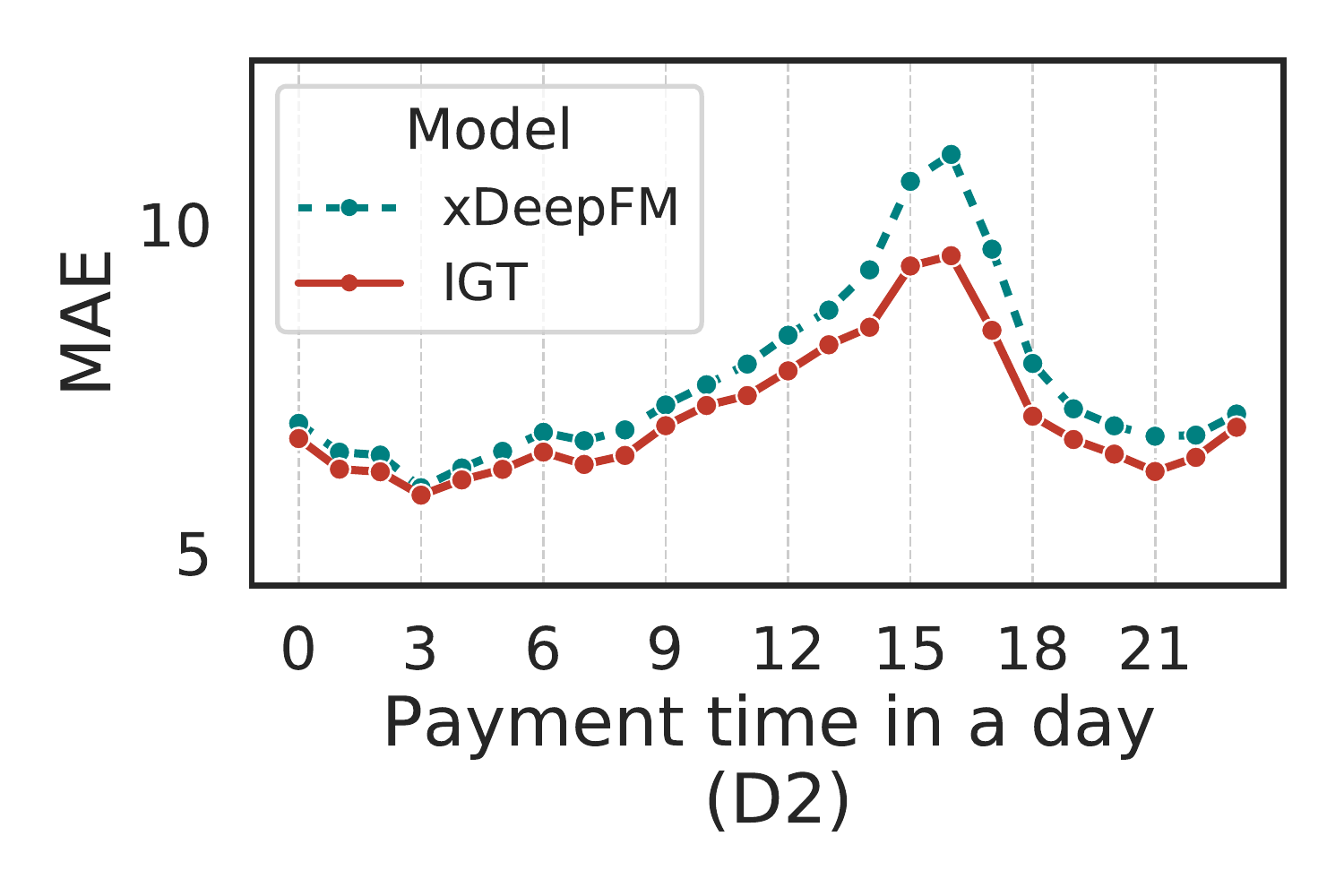}
	\end{subfigure}
	\caption{ETA prediction performance with regard to payment time in a day.}
	\label{fig:intro_per}
\end{figure}

\textbf{Payment time.}
We plot the ETA prediction performance of IGT and xDeepFM with regard to payment time in a day in Figure~\ref{fig:intro_per}, from which we can observe that our proposed model always shows better performance against the best baseline method xDeepFM at every payment time.
Obviously, the performance of both models exhibits certain fluctuations over payment time. 
When comparing Figure~\ref{fig:intro_per} with Figure~\ref{fig:time_dis} and Figure~\ref{fig:entropy}, we find the entropy of the delivery time on orders is highly correlated with the prediction results. 
More orders for training may not improve the ETA prediction performance of the models.
It is rational because larger entropy makes the models unable to estimate the delivery time accurately.
Both models show large prediction error around 15:00 on the two datasets.
The reason is that courier companies usually dispatch the collected packages to the next station around 15:00, products ordered after this time need to wait for the next dispatching slot.
In both datasets, our proposed IGT shows marginal improvements over xDeepFM around the peaks.

\begin{figure}[!h]
	\centering
	\includegraphics[trim=13 13 13 13, clip, width=0.475\textwidth]{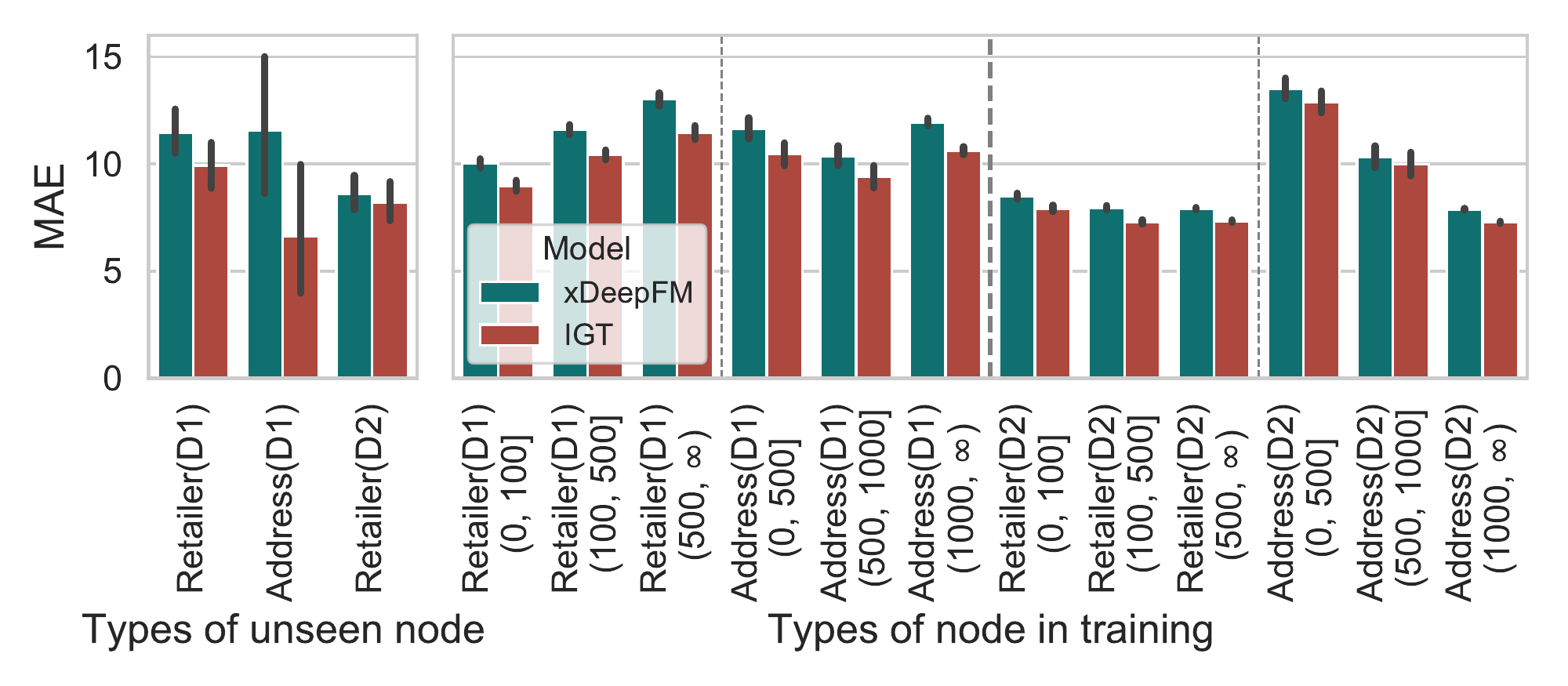}
	\caption{Inductive and transductive learning of IGT with regard to nodes with different number of historical orders.}
	\label{fig:inductive}
	\vspace{-10pt}
\end{figure}

\textbf{Retailer and address.}
As both datasets have orders with unseen retailers and addresses, we can evaluate the performance of IGT under inductive settings.
We take data binning to reduce the cardinality of the number of orders with regard to retailer and address.
To be specific, we divide the number of orders $N$ received by one retailer into groups: unseen ($N=0$), small ($0<N\leq 100$), medium ($100<N\leq 500$) and large ($500 < N$).
Similarly, we divide the number of orders $N$ associated with sender addresses into groups: unseen ($N=0$), small ($0<N\leq 500$), medium ($500<N\leq 1000$) and large ($1000 < N$). 
We plot the prediction performance on groups with a size larger than 50 in Figure~\ref{fig:inductive}.
In inductive learning setting, IGT shows better prediction performance than xDeepFM on both datasets under retailer and address.
Note that IGT reduces the MAE value by half of xDeepFM on unseen addresses of dataset D1.
Although the number of orders on unseen addresses is relatively small as the prediction values show higher variance, we can see IGT is superior to xDeepFM in inductive learning from the left part of Figure~\ref{fig:inductive}.
The right part of Figure~\ref{fig:inductive} shows that both models do not always learn well with the accumulation of historical orders from the same retailer or address.
Especially on dataset D1, larger number of orders from retailers degrades the performance of both models.
Compared with the retailer perspective, the accumulation of orders on address shows continuous improvements from both models on the two datasets.

\section{Online Experiment}
To further evaluate the effectiveness of IGT, we evaluate its performance online.
We evaluate the performance of IGT on the e-commerce platform for one week.
We split out the entire user requests by 1/10 to test our model.
The MAE, MAPE and MARE are 9.86 hours, 14.36\% and 15.83\%, respectively.
Compared with Table~\ref{tab:perform}, the national level evaluation performance is between that of D1 and D2. 
The results indicate cities with a larger population usually have lower MAE on ETA prediction.
Compared with the original served model, our model can reduce the average predictive error (\ie MAE) by 15\%.

\section{Conclusion and Future Work}
In this paper, we propose an inductive graph transformer (IGT) to estimate the package delivery time.
We use a decoupled architecture to capture both the high-order interactions of orders via a temporal and heterogeneous GCN and the interactions of elements in an order through a transformer-based model.
The information of unseen nodes can be learnt by propagating messages from neighboring nodes in our model. Hence, both structural information and raw features can be utilized in the transformer-based model for accurate ETA prediction.
In our model, IGT, we simplify GCN by removing both its non-linear layer and the transformation matrix. This allows industrial scenarios to apply IGT on large-scale datasets.
Experimental results on two large-scale offline logistics datasets demonstrate the effectiveness of IGT in both inductive and transductive learning. 
We further evaluate the performance of the model in a national level for online test.
IGT is promising in enhancing the user experiences of billions of customers in deployment on e-commerce platforms. 
For future work, we will focus on providing confidence intervals on ETA prediction instead of a point estimation.

%% file: main-cam.bbl

\begin{thebibliography}{43}


\ifx \showCODEN    \undefined \def \showCODEN     #1{\unskip}     \fi
\ifx \showDOI      \undefined \def \showDOI       #1{#1}\fi
\ifx \showISBNx    \undefined \def \showISBNx     #1{\unskip}     \fi
\ifx \showISBNxiii \undefined \def \showISBNxiii  #1{\unskip}     \fi
\ifx \showISSN     \undefined \def \showISSN      #1{\unskip}     \fi
\ifx \showLCCN     \undefined \def \showLCCN      #1{\unskip}     \fi
\ifx \shownote     \undefined \def \shownote      #1{#1}          \fi
\ifx \showarticletitle \undefined \def \showarticletitle #1{#1}   \fi
\ifx \showURL      \undefined \def \showURL       {\relax}        \fi
\providecommand\bibfield[2]{#2}
\providecommand\bibinfo[2]{#2}
\providecommand\natexlab[1]{#1}
\providecommand\showeprint[2][]{arXiv:#2}

\bibitem[Chen and Guestrin(2016)]%
        {chen2016xgboost}
\bibfield{author}{\bibinfo{person}{Tianqi Chen} {and} \bibinfo{person}{Carlos
  Guestrin}.} \bibinfo{year}{2016}\natexlab{}.
\newblock \showarticletitle{Xgboost: A scalable tree boosting system}. In
  \bibinfo{booktitle}{\emph{Proceedings of the 22nd acm sigkdd international
  conference on knowledge discovery and data mining}}.
  \bibinfo{pages}{785--794}.
\newblock


\bibitem[Cheng et~al\mbox{.}(2020)]%
        {cheng2020skeleton}
\bibfield{author}{\bibinfo{person}{Ke Cheng}, \bibinfo{person}{Yifan Zhang},
  \bibinfo{person}{Xiangyu He}, \bibinfo{person}{Weihan Chen},
  \bibinfo{person}{Jian Cheng}, {and} \bibinfo{person}{Hanqing Lu}.}
  \bibinfo{year}{2020}\natexlab{}.
\newblock \showarticletitle{Skeleton-based action recognition with shift graph
  convolutional network}. In \bibinfo{booktitle}{\emph{Proceedings of the
  IEEE/CVF Conference on Computer Vision and Pattern Recognition}}.
  \bibinfo{pages}{183--192}.
\newblock


\bibitem[Cho et~al\mbox{.}(2014)]%
        {cho2014learning}
\bibfield{author}{\bibinfo{person}{Kyunghyun Cho}, \bibinfo{person}{Bart
  Van~Merri{\"e}nboer}, \bibinfo{person}{Caglar Gulcehre},
  \bibinfo{person}{Dzmitry Bahdanau}, \bibinfo{person}{Fethi Bougares},
  \bibinfo{person}{Holger Schwenk}, {and} \bibinfo{person}{Yoshua Bengio}.}
  \bibinfo{year}{2014}\natexlab{}.
\newblock \showarticletitle{Learning phrase representations using RNN
  encoder-decoder for statistical machine translation}. In
  \bibinfo{booktitle}{\emph{Proceedings of the 2014 Conference on Empirical
  Methods in Natural Language Processing ({EMNLP})}}.
  \bibinfo{pages}{1724--1734}.
\newblock


\bibitem[Corso et~al\mbox{.}(2020)]%
        {corso2020principal}
\bibfield{author}{\bibinfo{person}{Gabriele Corso}, \bibinfo{person}{Luca
  Cavalleri}, \bibinfo{person}{Dominique Beaini}, \bibinfo{person}{Pietro
  Li{\`o}}, {and} \bibinfo{person}{Petar Veli{\v{c}}kovi{\'c}}.}
  \bibinfo{year}{2020}\natexlab{}.
\newblock \showarticletitle{Principal neighbourhood aggregation for graph
  nets}. In \bibinfo{booktitle}{\emph{Proceedings of the 34th International
  Conference on Neural Information Processing Systems}}.
\newblock


\bibitem[Cui et~al\mbox{.}(2020)]%
        {cui2020sooner}
\bibfield{author}{\bibinfo{person}{Ruomeng Cui}, \bibinfo{person}{Zhikun Lu},
  \bibinfo{person}{Tianshu Sun}, {and} \bibinfo{person}{Joseph Golden}.}
  \bibinfo{year}{2020}\natexlab{}.
\newblock \showarticletitle{Sooner or later? Promising delivery speed in online
  retail}.
\newblock \bibinfo{journal}{\emph{Promising Delivery Speed in Online Retail}}
  (\bibinfo{year}{2020}).
\newblock


\bibitem[Defferrard et~al\mbox{.}(2016)]%
        {defferrard2016convolutional}
\bibfield{author}{\bibinfo{person}{Micha{\"e}l Defferrard},
  \bibinfo{person}{Xavier Bresson}, {and} \bibinfo{person}{Pierre
  Vandergheynst}.} \bibinfo{year}{2016}\natexlab{}.
\newblock \showarticletitle{Convolutional neural networks on graphs with fast
  localized spectral filtering}.
\newblock \bibinfo{journal}{\emph{Advances in neural information processing
  systems}}  \bibinfo{volume}{29} (\bibinfo{year}{2016}).
\newblock


\bibitem[Devlin et~al\mbox{.}(2019)]%
        {devlin2019bert}
\bibfield{author}{\bibinfo{person}{Jacob Devlin}, \bibinfo{person}{Ming-Wei
  Chang}, \bibinfo{person}{Kenton Lee}, {and} \bibinfo{person}{Kristina
  Toutanova}.} \bibinfo{year}{2019}\natexlab{}.
\newblock \showarticletitle{Bert: Pre-training of deep bidirectional
  transformers for language understanding}. In
  \bibinfo{booktitle}{\emph{Conference of the North American Chapter of the
  Association for Computational Linguistics}}.
\newblock


\bibitem[Doole et~al\mbox{.}(2020)]%
        {doole2020estimation}
\bibfield{author}{\bibinfo{person}{Malik Doole}, \bibinfo{person}{Joost
  Ellerbroek}, {and} \bibinfo{person}{Jacco Hoekstra}.}
  \bibinfo{year}{2020}\natexlab{}.
\newblock \showarticletitle{Estimation of traffic density from drone-based
  delivery in very low level urban airspace}.
\newblock \bibinfo{journal}{\emph{Journal of Air Transport Management}}
  \bibinfo{volume}{88} (\bibinfo{year}{2020}), \bibinfo{pages}{101862}.
\newblock


\bibitem[Dosovitskiy et~al\mbox{.}(2021)]%
        {dosovitskiy2021image}
\bibfield{author}{\bibinfo{person}{Alexey Dosovitskiy}, \bibinfo{person}{Lucas
  Beyer}, \bibinfo{person}{Alexander Kolesnikov}, \bibinfo{person}{Dirk
  Weissenborn}, \bibinfo{person}{Xiaohua Zhai}, \bibinfo{person}{Thomas
  Unterthiner}, \bibinfo{person}{Mostafa Dehghani}, \bibinfo{person}{Matthias
  Minderer}, \bibinfo{person}{Georg Heigold}, \bibinfo{person}{Sylvain Gelly},
  {et~al\mbox{.}}} \bibinfo{year}{2021}\natexlab{}.
\newblock \showarticletitle{An image is worth 16x16 words: Transformers for
  image recognition at scale}. In \bibinfo{booktitle}{\emph{International
  Conference on Learning Representations}}.
\newblock


\bibitem[Dwivedi and Bresson(2020)]%
        {dwivedi2020generalization}
\bibfield{author}{\bibinfo{person}{Vijay~Prakash Dwivedi} {and}
  \bibinfo{person}{Xavier Bresson}.} \bibinfo{year}{2020}\natexlab{}.
\newblock \showarticletitle{A generalization of transformer networks to
  graphs}.
\newblock \bibinfo{journal}{\emph{arXiv preprint arXiv:2012.09699}}
  (\bibinfo{year}{2020}).
\newblock


\bibitem[Fang et~al\mbox{.}(2020)]%
        {fang2020constgat}
\bibfield{author}{\bibinfo{person}{Xiaomin Fang}, \bibinfo{person}{Jizhou
  Huang}, \bibinfo{person}{Fan Wang}, \bibinfo{person}{Lingke Zeng},
  \bibinfo{person}{Haijin Liang}, {and} \bibinfo{person}{Haifeng Wang}.}
  \bibinfo{year}{2020}\natexlab{}.
\newblock \showarticletitle{Constgat: Contextual spatial-temporal graph
  attention network for travel time estimation at baidu maps}. In
  \bibinfo{booktitle}{\emph{Proceedings of the 26th ACM SIGKDD International
  Conference on Knowledge Discovery \& Data Mining}}.
  \bibinfo{pages}{2697--2705}.
\newblock


\bibitem[Fu et~al\mbox{.}(2020)]%
        {fu2020compacteta}
\bibfield{author}{\bibinfo{person}{Kun Fu}, \bibinfo{person}{Fanlin Meng},
  \bibinfo{person}{Jieping Ye}, {and} \bibinfo{person}{Zheng Wang}.}
  \bibinfo{year}{2020}\natexlab{}.
\newblock \showarticletitle{Compacteta: A fast inference system for travel time
  prediction}. In \bibinfo{booktitle}{\emph{Proceedings of the 26th ACM SIGKDD
  International Conference on Knowledge Discovery \& Data Mining}}.
  \bibinfo{pages}{3337--3345}.
\newblock


\bibitem[Glorot and Bengio(2010)]%
        {glorot2010understanding}
\bibfield{author}{\bibinfo{person}{Xavier Glorot} {and} \bibinfo{person}{Yoshua
  Bengio}.} \bibinfo{year}{2010}\natexlab{}.
\newblock \showarticletitle{Understanding the difficulty of training deep
  feedforward neural networks}. In \bibinfo{booktitle}{\emph{Proceedings of the
  thirteenth international conference on artificial intelligence and
  statistics}}. JMLR Workshop and Conference Proceedings,
  \bibinfo{pages}{249--256}.
\newblock


\bibitem[Hamilton et~al\mbox{.}(2017)]%
        {hamilton2017inductive}
\bibfield{author}{\bibinfo{person}{William~L Hamilton}, \bibinfo{person}{Rex
  Ying}, {and} \bibinfo{person}{Jure Leskovec}.}
  \bibinfo{year}{2017}\natexlab{}.
\newblock \showarticletitle{Inductive representation learning on large graphs}.
  In \bibinfo{booktitle}{\emph{Proceedings of the 31st International Conference
  on Neural Information Processing Systems}}. \bibinfo{pages}{1025--1035}.
\newblock


\bibitem[Hong et~al\mbox{.}(2020)]%
        {hong2020heteta}
\bibfield{author}{\bibinfo{person}{Huiting Hong}, \bibinfo{person}{Yucheng
  Lin}, \bibinfo{person}{Xiaoqing Yang}, \bibinfo{person}{Zang Li},
  \bibinfo{person}{Kung Fu}, \bibinfo{person}{Zheng Wang},
  \bibinfo{person}{Xiaohu Qie}, {and} \bibinfo{person}{Jieping Ye}.}
  \bibinfo{year}{2020}\natexlab{}.
\newblock \showarticletitle{HetETA: Heterogeneous Information Network Embedding
  for Estimating Time of Arrival}. In \bibinfo{booktitle}{\emph{Proceedings of
  the 26th ACM SIGKDD International Conference on Knowledge Discovery \& Data
  Mining}}. \bibinfo{publisher}{Association for Computing Machinery},
  \bibinfo{address}{New York, NY, USA}, \bibinfo{pages}{2444--2454}.
\newblock


\bibitem[Hu et~al\mbox{.}(2020b)]%
        {hu2020stochastic}
\bibfield{author}{\bibinfo{person}{Jilin Hu}, \bibinfo{person}{Bin Yang},
  \bibinfo{person}{Chenjuan Guo}, \bibinfo{person}{Christian~S Jensen}, {and}
  \bibinfo{person}{Hui Xiong}.} \bibinfo{year}{2020}\natexlab{b}.
\newblock \showarticletitle{Stochastic origin-destination matrix forecasting
  using dual-stage graph convolutional, recurrent neural networks}. In
  \bibinfo{booktitle}{\emph{2020 IEEE 36th International Conference on Data
  Engineering (ICDE)}}. IEEE, \bibinfo{pages}{1417--1428}.
\newblock


\bibitem[Hu et~al\mbox{.}(2020a)]%
        {hu2020heterogeneous}
\bibfield{author}{\bibinfo{person}{Ziniu Hu}, \bibinfo{person}{Yuxiao Dong},
  \bibinfo{person}{Kuansan Wang}, {and} \bibinfo{person}{Yizhou Sun}.}
  \bibinfo{year}{2020}\natexlab{a}.
\newblock \showarticletitle{Heterogeneous graph transformer}. In
  \bibinfo{booktitle}{\emph{Proceedings of The Web Conference 2020}}.
  \bibinfo{pages}{2704--2710}.
\newblock


\bibitem[Kazemi et~al\mbox{.}(2020)]%
        {kazemi2020representation}
\bibfield{author}{\bibinfo{person}{Seyed~Mehran Kazemi},
  \bibinfo{person}{Rishab Goel}, \bibinfo{person}{Kshitij Jain},
  \bibinfo{person}{Ivan Kobyzev}, \bibinfo{person}{Akshay Sethi},
  \bibinfo{person}{Peter Forsyth}, {and} \bibinfo{person}{Pascal Poupart}.}
  \bibinfo{year}{2020}\natexlab{}.
\newblock \showarticletitle{Representation Learning for Dynamic Graphs: A
  Survey.}
\newblock \bibinfo{journal}{\emph{Journal of Machine Learning Research}}
  \bibinfo{volume}{21}, \bibinfo{number}{70} (\bibinfo{year}{2020}),
  \bibinfo{pages}{1--73}.
\newblock


\bibitem[Kingma and Ba(2015)]%
        {kingma2015adam}
\bibfield{author}{\bibinfo{person}{Diederik~P Kingma} {and}
  \bibinfo{person}{Jimmy Ba}.} \bibinfo{year}{2015}\natexlab{}.
\newblock \showarticletitle{Adam: A method for stochastic optimization}.
\newblock \bibinfo{journal}{\emph{International Conference on Learning
  Representations}}.
\newblock


\bibitem[Kipf and Welling(2017)]%
        {kipf2017semi}
\bibfield{author}{\bibinfo{person}{Thomas~N Kipf} {and} \bibinfo{person}{Max
  Welling}.} \bibinfo{year}{2017}\natexlab{}.
\newblock \showarticletitle{Semi-supervised classification with graph
  convolutional networks}. In \bibinfo{booktitle}{\emph{International
  Conference on Learning Representations}}.
\newblock


\bibitem[Li et~al\mbox{.}(2018b)]%
        {li2018deeper}
\bibfield{author}{\bibinfo{person}{Qimai Li}, \bibinfo{person}{Zhichao Han},
  {and} \bibinfo{person}{Xiao-Ming Wu}.} \bibinfo{year}{2018}\natexlab{b}.
\newblock \showarticletitle{Deeper insights into graph convolutional networks
  for semi-supervised learning}. In \bibinfo{booktitle}{\emph{Thirty-Second
  AAAI conference on artificial intelligence}}.
\newblock


\bibitem[Li et~al\mbox{.}(2018a)]%
        {li2018multi}
\bibfield{author}{\bibinfo{person}{Yaguang Li}, \bibinfo{person}{Kun Fu},
  \bibinfo{person}{Zheng Wang}, \bibinfo{person}{Cyrus Shahabi},
  \bibinfo{person}{Jieping Ye}, {and} \bibinfo{person}{Yan Liu}.}
  \bibinfo{year}{2018}\natexlab{a}.
\newblock \showarticletitle{Multi-Task Representation Learning for Travel Time
  Estimation}. In \bibinfo{booktitle}{\emph{Proceedings of the 24th ACM SIGKDD
  International Conference on Knowledge Discovery \& Data Mining}}.
  \bibinfo{publisher}{Association for Computing Machinery},
  \bibinfo{address}{New York, NY, USA}, \bibinfo{pages}{1695--1704}.
\newblock


\bibitem[Lian et~al\mbox{.}(2018)]%
        {lian2018xdeepfm}
\bibfield{author}{\bibinfo{person}{Jianxun Lian}, \bibinfo{person}{Xiaohuan
  Zhou}, \bibinfo{person}{Fuzheng Zhang}, \bibinfo{person}{Zhongxia Chen},
  \bibinfo{person}{Xing Xie}, {and} \bibinfo{person}{Guangzhong Sun}.}
  \bibinfo{year}{2018}\natexlab{}.
\newblock \showarticletitle{xdeepfm: Combining explicit and implicit feature
  interactions for recommender systems}. In
  \bibinfo{booktitle}{\emph{Proceedings of the 24th ACM SIGKDD International
  Conference on Knowledge Discovery \& Data Mining}}.
  \bibinfo{pages}{1754--1763}.
\newblock


\bibitem[Paszke et~al\mbox{.}(2019)]%
        {paszke2019pytorch}
\bibfield{author}{\bibinfo{person}{Adam Paszke}, \bibinfo{person}{Sam Gross},
  \bibinfo{person}{Francisco Massa}, \bibinfo{person}{Adam Lerer},
  \bibinfo{person}{James Bradbury}, \bibinfo{person}{Gregory Chanan},
  \bibinfo{person}{Trevor Killeen}, \bibinfo{person}{Zeming Lin},
  \bibinfo{person}{Natalia Gimelshein}, \bibinfo{person}{Luca Antiga},
  {et~al\mbox{.}}} \bibinfo{year}{2019}\natexlab{}.
\newblock \showarticletitle{Pytorch: An imperative style, high-performance deep
  learning library}.
\newblock \bibinfo{journal}{\emph{Advances in neural information processing
  systems}}  \bibinfo{volume}{32} (\bibinfo{year}{2019}),
  \bibinfo{pages}{8026--8037}.
\newblock


\bibitem[Pedregosa et~al\mbox{.}(2011)]%
        {pedregosa2011scikit}
\bibfield{author}{\bibinfo{person}{Fabian Pedregosa}, \bibinfo{person}{Ga{\"e}l
  Varoquaux}, \bibinfo{person}{Alexandre Gramfort}, \bibinfo{person}{Vincent
  Michel}, \bibinfo{person}{Bertrand Thirion}, \bibinfo{person}{Olivier
  Grisel}, \bibinfo{person}{Mathieu Blondel}, \bibinfo{person}{Peter
  Prettenhofer}, \bibinfo{person}{Ron Weiss}, \bibinfo{person}{Vincent
  Dubourg}, {et~al\mbox{.}}} \bibinfo{year}{2011}\natexlab{}.
\newblock \showarticletitle{Scikit-learn: Machine learning in Python}.
\newblock \bibinfo{journal}{\emph{the Journal of machine Learning research}}
  \bibinfo{volume}{12} (\bibinfo{year}{2011}), \bibinfo{pages}{2825--2830}.
\newblock


\bibitem[Ragesh et~al\mbox{.}(2021)]%
        {ragesh2021hetegcn}
\bibfield{author}{\bibinfo{person}{Rahul Ragesh}, \bibinfo{person}{Sundararajan
  Sellamanickam}, \bibinfo{person}{Arun Iyer}, \bibinfo{person}{Ramakrishna
  Bairi}, {and} \bibinfo{person}{Vijay Lingam}.}
  \bibinfo{year}{2021}\natexlab{}.
\newblock \showarticletitle{Hetegcn: heterogeneous graph convolutional networks
  for text classification}. In \bibinfo{booktitle}{\emph{Proceedings of the
  14th ACM International Conference on Web Search and Data Mining}}.
  \bibinfo{pages}{860--868}.
\newblock


\bibitem[Su et~al\mbox{.}(2020)]%
        {su2020characterizing}
\bibfield{author}{\bibinfo{person}{Yuehuan Su}, \bibinfo{person}{Huabo Duan},
  \bibinfo{person}{Zinuo Wang}, \bibinfo{person}{Guanghan Song},
  \bibinfo{person}{Peng Kang}, {and} \bibinfo{person}{Dongjie Chen}.}
  \bibinfo{year}{2020}\natexlab{}.
\newblock \showarticletitle{Characterizing the environmental impact of
  packaging materials for express delivery via life cycle assessment}.
\newblock \bibinfo{journal}{\emph{Journal of Cleaner Production}}
  \bibinfo{volume}{274} (\bibinfo{year}{2020}), \bibinfo{pages}{122961}.
\newblock


\bibitem[Vaswani et~al\mbox{.}(2017)]%
        {vaswani2017attention}
\bibfield{author}{\bibinfo{person}{Ashish Vaswani}, \bibinfo{person}{Noam
  Shazeer}, \bibinfo{person}{Niki Parmar}, \bibinfo{person}{Jakob Uszkoreit},
  \bibinfo{person}{Llion Jones}, \bibinfo{person}{Aidan~N Gomez},
  \bibinfo{person}{{\L}ukasz Kaiser}, {and} \bibinfo{person}{Illia
  Polosukhin}.} \bibinfo{year}{2017}\natexlab{}.
\newblock \showarticletitle{Attention is all you need}. In
  \bibinfo{booktitle}{\emph{Advances in neural information processing
  systems}}. \bibinfo{pages}{5998--6008}.
\newblock


\bibitem[Veli{\v{c}}kovi{\'c} et~al\mbox{.}(2018)]%
        {velivckovic2018graph}
\bibfield{author}{\bibinfo{person}{Petar Veli{\v{c}}kovi{\'c}},
  \bibinfo{person}{Guillem Cucurull}, \bibinfo{person}{Arantxa Casanova},
  \bibinfo{person}{Adriana Romero}, \bibinfo{person}{Pietro Lio}, {and}
  \bibinfo{person}{Yoshua Bengio}.} \bibinfo{year}{2018}\natexlab{}.
\newblock \showarticletitle{Graph attention networks}. In
  \bibinfo{booktitle}{\emph{Proceedings of the Sixth International Conference
  on Learning Representations}}.
\newblock


\bibitem[Vladimir(1996)]%
        {vladimir1996electronic}
\bibfield{author}{\bibinfo{person}{Zwass Vladimir}.}
  \bibinfo{year}{1996}\natexlab{}.
\newblock \showarticletitle{Electronic commerce: structures and issues}.
\newblock \bibinfo{journal}{\emph{International journal of electronic
  commerce}} \bibinfo{volume}{1}, \bibinfo{number}{1} (\bibinfo{year}{1996}),
  \bibinfo{pages}{3--23}.
\newblock


\bibitem[Wang et~al\mbox{.}(2019a)]%
        {wang2019asimple}
\bibfield{author}{\bibinfo{person}{Hongjian Wang}, \bibinfo{person}{Xianfeng
  Tang}, \bibinfo{person}{Yu-Hsuan Kuo}, \bibinfo{person}{Daniel Kifer}, {and}
  \bibinfo{person}{Zhenhui Li}.} \bibinfo{year}{2019}\natexlab{a}.
\newblock \showarticletitle{A Simple Baseline for Travel Time Estimation Using
  Large-Scale Trip Data}.
\newblock \bibinfo{journal}{\emph{ACM Trans. Intell. Syst. Technol.}}
  \bibinfo{volume}{10}, \bibinfo{number}{2} (\bibinfo{year}{2019}).
\newblock
\showISSN{2157--6904}


\bibitem[Wang et~al\mbox{.}(2019b)]%
        {wang2019fdgars}
\bibfield{author}{\bibinfo{person}{Jianyu Wang}, \bibinfo{person}{Rui Wen},
  \bibinfo{person}{Chunming Wu}, \bibinfo{person}{Yu Huang}, {and}
  \bibinfo{person}{Jian Xion}.} \bibinfo{year}{2019}\natexlab{b}.
\newblock \showarticletitle{Fdgars: Fraudster detection via graph convolutional
  networks in online app review system}. In \bibinfo{booktitle}{\emph{Companion
  Proceedings of The 2019 World Wide Web Conference}}.
  \bibinfo{pages}{310--316}.
\newblock


\bibitem[Wang et~al\mbox{.}(2018)]%
        {wang2018learning}
\bibfield{author}{\bibinfo{person}{Zheng Wang}, \bibinfo{person}{Kun Fu}, {and}
  \bibinfo{person}{Jieping Ye}.} \bibinfo{year}{2018}\natexlab{}.
\newblock \showarticletitle{Learning to estimate the travel time}. In
  \bibinfo{booktitle}{\emph{Proceedings of the 24th ACM SIGKDD International
  Conference on Knowledge Discovery \& Data Mining}}.
  \bibinfo{pages}{858--866}.
\newblock


\bibitem[Wu et~al\mbox{.}(2019)]%
        {wu2019simplifying}
\bibfield{author}{\bibinfo{person}{Felix Wu}, \bibinfo{person}{Amauri Souza},
  \bibinfo{person}{Tianyi Zhang}, \bibinfo{person}{Christopher Fifty},
  \bibinfo{person}{Tao Yu}, {and} \bibinfo{person}{Kilian Weinberger}.}
  \bibinfo{year}{2019}\natexlab{}.
\newblock \showarticletitle{Simplifying graph convolutional networks}. In
  \bibinfo{booktitle}{\emph{International conference on machine learning}}.
  PMLR, \bibinfo{pages}{6861--6871}.
\newblock


\bibitem[Wu and Wu(2019)]%
        {wu2019deepeta}
\bibfield{author}{\bibinfo{person}{Fan Wu} {and} \bibinfo{person}{Lixia Wu}.}
  \bibinfo{year}{2019}\natexlab{}.
\newblock \showarticletitle{Deepeta: A spatial-temporal sequential neural
  network model for estimating time of arrival in package delivery system}. In
  \bibinfo{booktitle}{\emph{Proceedings of the AAAI Conference on Artificial
  Intelligence}}, Vol.~\bibinfo{volume}{33}. \bibinfo{pages}{774--781}.
\newblock


\bibitem[Yao et~al\mbox{.}(2019)]%
        {yao2019graph}
\bibfield{author}{\bibinfo{person}{Liang Yao}, \bibinfo{person}{Chengsheng
  Mao}, {and} \bibinfo{person}{Yuan Luo}.} \bibinfo{year}{2019}\natexlab{}.
\newblock \showarticletitle{Graph convolutional networks for text
  classification}. In \bibinfo{booktitle}{\emph{Proceedings of the AAAI
  conference on artificial intelligence}}, Vol.~\bibinfo{volume}{33}.
  \bibinfo{pages}{7370--7377}.
\newblock


\bibitem[Yun et~al\mbox{.}(2019)]%
        {yun2019graph}
\bibfield{author}{\bibinfo{person}{Seongjun Yun}, \bibinfo{person}{Minbyul
  Jeong}, \bibinfo{person}{Raehyun Kim}, \bibinfo{person}{Jaewoo Kang}, {and}
  \bibinfo{person}{Hyunwoo~J Kim}.} \bibinfo{year}{2019}\natexlab{}.
\newblock \showarticletitle{Graph transformer networks}.
\newblock \bibinfo{journal}{\emph{Advances in Neural Information Processing
  Systems}}  \bibinfo{volume}{32} (\bibinfo{year}{2019}),
  \bibinfo{pages}{11983--11993}.
\newblock


\bibitem[Zhou et~al\mbox{.}(2020)]%
        {zhou2020data}
\bibfield{author}{\bibinfo{person}{Dawei Zhou}, \bibinfo{person}{Lecheng
  Zheng}, \bibinfo{person}{Jiawei Han}, {and} \bibinfo{person}{Jingrui He}.}
  \bibinfo{year}{2020}\natexlab{}.
\newblock \showarticletitle{A data-driven graph generative model for temporal
  interaction networks}. In \bibinfo{booktitle}{\emph{Proceedings of the 26th
  ACM SIGKDD International Conference on Knowledge Discovery \& Data Mining}}.
  \bibinfo{pages}{401--411}.
\newblock


\bibitem[Zhou et~al\mbox{.}(2022a)]%
        {zhou2022layer}
\bibfield{author}{\bibinfo{person}{Xin Zhou}, \bibinfo{person}{Donghui Lin},
  \bibinfo{person}{Yong Liu}, {and} \bibinfo{person}{Chunyan Miao}.}
  \bibinfo{year}{2022}\natexlab{a}.
\newblock \showarticletitle{Layer-refined Graph Convolutional Networks for
  Recommendation}.
\newblock \bibinfo{journal}{\emph{arXiv preprint arXiv:2207.11088}}
  (\bibinfo{year}{2022}).
\newblock


\bibitem[Zhou et~al\mbox{.}(2021)]%
        {zhou2021selfcf}
\bibfield{author}{\bibinfo{person}{Xin Zhou}, \bibinfo{person}{Aixin Sun},
  \bibinfo{person}{Yong Liu}, \bibinfo{person}{Jie Zhang}, {and}
  \bibinfo{person}{Chunyan Miao}.} \bibinfo{year}{2021}\natexlab{}.
\newblock \showarticletitle{SelfCF: A Simple Framework for Self-supervised
  Collaborative Filtering}.
\newblock \bibinfo{journal}{\emph{arXiv preprint arXiv:2107.03019}}
  (\bibinfo{year}{2021}).
\newblock


\bibitem[Zhou et~al\mbox{.}(2022b)]%
        {zhou2022bootstrap}
\bibfield{author}{\bibinfo{person}{Xin Zhou}, \bibinfo{person}{Hongyu Zhou},
  \bibinfo{person}{Yong Liu}, \bibinfo{person}{Zhiwei Zeng},
  \bibinfo{person}{Chunyan Miao}, \bibinfo{person}{Pengwei Wang},
  \bibinfo{person}{Yuan You}, {and} \bibinfo{person}{Feijun Jiang}.}
  \bibinfo{year}{2022}\natexlab{b}.
\newblock \showarticletitle{Bootstrap Latent Representations for Multi-modal
  Recommendation}.
\newblock \bibinfo{journal}{\emph{arXiv preprint arXiv:2207.05969}}
  (\bibinfo{year}{2022}).
\newblock


\bibitem[Zhu et~al\mbox{.}(2020b)]%
        {zhu2020order}
\bibfield{author}{\bibinfo{person}{Lin Zhu}, \bibinfo{person}{Wei Yu},
  \bibinfo{person}{Kairong Zhou}, \bibinfo{person}{Xing Wang},
  \bibinfo{person}{Wenxing Feng}, \bibinfo{person}{Pengyu Wang},
  \bibinfo{person}{Ning Chen}, {and} \bibinfo{person}{Pei Lee}.}
  \bibinfo{year}{2020}\natexlab{b}.
\newblock \showarticletitle{Order Fulfillment Cycle Time Estimation for
  On-Demand Food Delivery}. In \bibinfo{booktitle}{\emph{Proceedings of the
  26th ACM SIGKDD International Conference on Knowledge Discovery \& Data
  Mining}}. \bibinfo{pages}{2571--2580}.
\newblock


\bibitem[Zhu et~al\mbox{.}(2020a)]%
        {zhu2020hgcn}
\bibfield{author}{\bibinfo{person}{Zhihua Zhu}, \bibinfo{person}{Xinxin Fan},
  \bibinfo{person}{Xiaokai Chu}, {and} \bibinfo{person}{Jingping Bi}.}
  \bibinfo{year}{2020}\natexlab{a}.
\newblock \showarticletitle{Hgcn: A heterogeneous graph convolutional
  network-based deep learning model toward collective classification}. In
  \bibinfo{booktitle}{\emph{Proceedings of the 26th ACM SIGKDD International
  Conference on Knowledge Discovery \& Data Mining}}.
  \bibinfo{pages}{1161--1171}.
\newblock


\end{thebibliography}
